\documentclass[journal]{IEEEtran}
\usepackage{times}

\usepackage[numbers]{natbib}
\usepackage{multicol}
\usepackage{multirow}
\usepackage[bookmarks=true]{hyperref}
\usepackage{amsmath} 
\usepackage{bm}
\usepackage{gensymb}
\usepackage{graphicx}
\usepackage{subfig}
\usepackage{amsfonts}
\usepackage{algorithm}
\usepackage{algorithmicx}
\usepackage{algpseudocode}
\usepackage{flushend}
\usepackage{soul}
\usepackage{array}

\DeclareMathOperator*{\argmin}{arg\,min}
\newcolumntype{P}[1]{>{\centering\arraybackslash}p{#1}}
\algdef{SE}[DOWHILE]{Do}{doWhile}{\algorithmicdo}[1]{\algorithmicwhile\ #1}%

\newcommand\blfootnote[1]{%
  \begingroup
  \renewcommand\thefootnote{}\footnote{#1}%
  \addtocounter{footnote}{-1}%
  \endgroup
}

\newcommand{\remindtext}[2]{{{#2}}}
\newcommand{\addedtext}[2]{{{#2}}}
\newcommand{\modifiedtext}[2]{{{#2}}}
\newcommand{\deletedtext}[2]{{}}

\begin{document}

\title{Accurate Energetic Constraints for\\ Passive Grasp Stability Analysis}



\author{
  \authorblockN{
    Maximilian Haas-Heger\authorrefmark{1},
    Matei Ciocarlie\authorrefmark{2}\\}
  \authorblockA{
    \authorrefmark{1}email: m.haas@columbia.edu
    \authorrefmark{2}email: matei.ciocarlie@columbia.edu\\
  }
  \authorblockA{Department of Mechanical Engineering; Columbia University, New York, NY 10027}
}

%

\maketitle

\begin{abstract} 

Passive reaction effects in grasp stability analysis occur when the
contact forces and joint torques applied by a grasp change in response
to external disturbances applied to the grasped object. For example,
nonbackdrivable actuators (e.g. highly geared servos) will passively
resist external disturbances without an actively applied command; for
numerous robot hands using such motors, these effects can be highly
beneficial as they increase grasp resistance without requiring active
control. We introduce a grasp stability analysis method that can model
these effects, and, for a given grasp, distinguish between
disturbances that will be passively resisted and those that will
not. We find that, in order to achieve this, the grasp model
must include accurate energetic constraints. One way to achieve this
is to consider the Maximum Dissipation Principle (MDP), a part of the
Coulomb friction model that is rarely used in grasp stability
analysis. However, the MDP constraints are non-convex, and difficult
to solve efficiently. We thus introduce a convex relaxation method,
along with an algorithm that successively refines this relaxation
locally in order to obtain solutions to arbitrary accuracy
efficiently. Our resulting algorithm can determine if a grasp is
passively stable, solve for equilibrium contact forces and
compute optimal actuator commands for stability. Its implementation is publicly available
as part of the open-source GraspIt! simulator. \blfootnote{This work was supported in part by the National Science Foundation under CAREER Grant IIS-1551631 and by the Office of Naval Research under Grant N00014-16-1-2026. Digital Object Identifier 10.1109/TRO.2020.2974108}


\end{abstract}

\IEEEpeerreviewmaketitle

\section{INTRODUCTION}\label{introduction}

The analysis of the stability of a grasp is a foundational aspect of
multi-fingered robotic manipulation. Determining the ability of a grasp to
resist given disturbances, formulated as external wrenches applied to the
grasped object, is equivalent to computing the stability of a multi-body
system with frictional contacts under applied loads. Problems of this kind are
thus pervasive in grasp analysis and may be encountered in many other
scenarios that require simulation of general rigid bodies with frictional
contacts.

A major complication in the simulation of multi-body systems is the
accurate modeling of friction. Of particular interest to us is the
Maximum Dissipation Principle (MDP), which is a part of the Coulomb
friction model. This principle, which informally states that friction
attempts to dissipate as much energy as possible in the presence of
motion, imposes constraints that are both non-smooth and
non-convex~\cite{STEWART00}. \remindtext{7-2-1}{Existing exact formulations of the MDP
are NP-complete~\cite{PANG1996}, and therefore do not allow for
efficient computation of solutions. To avoid this, current grasp
models do not include the MDP at all, and therefore do not suffer from
the same complexity. In consequence, however, these grasp models
cannot capture what we call \textit{passive stability}.} As we argue in
this paper, in order to determine the passive stability of a grasp,
inclusion of the MDP in the friction model is imperative in order to
maintain energy conservation laws.

\begin{figure}[!t]
\centering
\includegraphics[width=1.0\columnwidth]{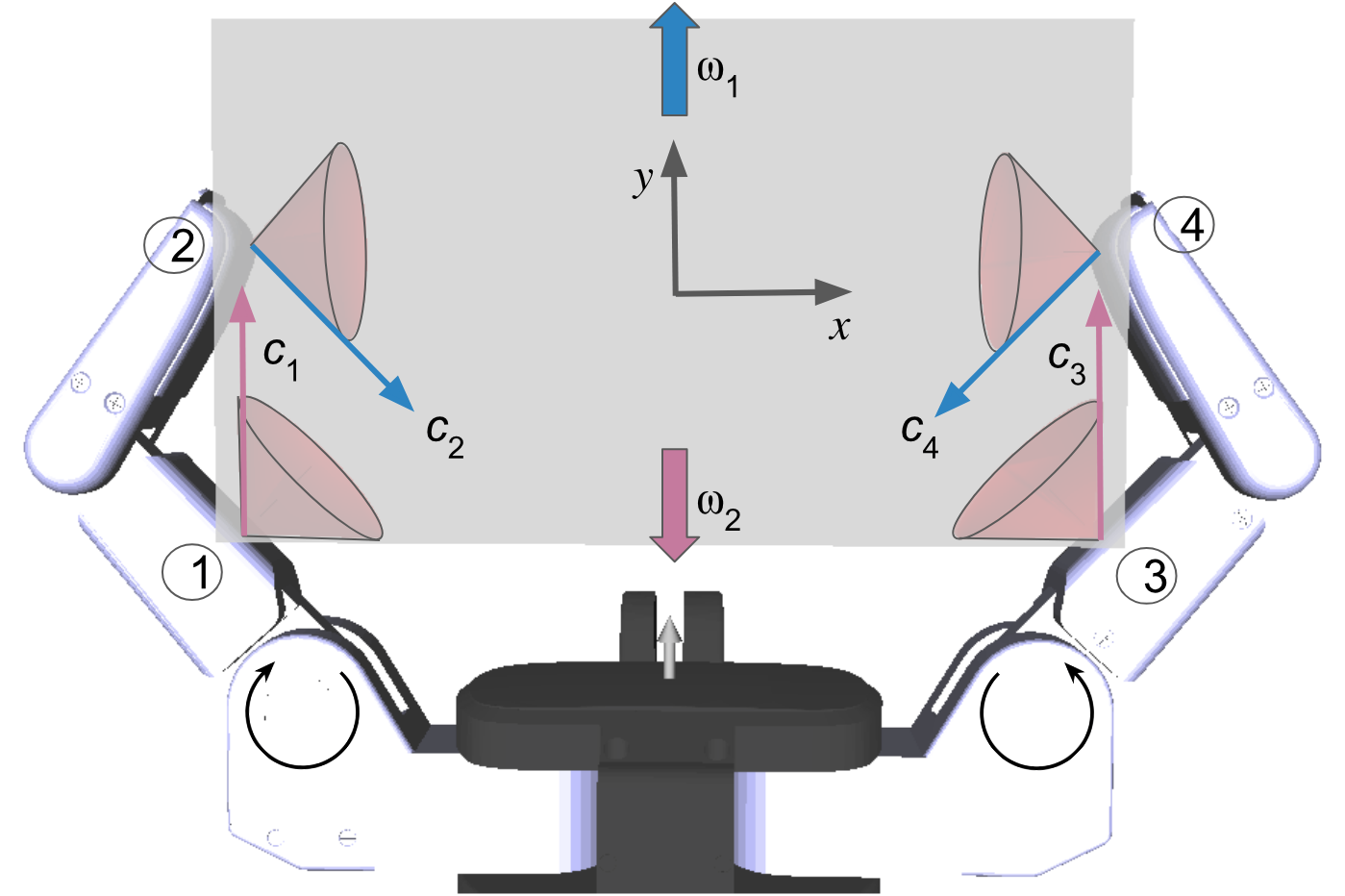}
\caption{A grasping scenario where a hand establishes multiple
  frictional contacts (numbered 1-4) with a target object. A
  disturbance pushing the object ``up'' (such as $\bm{w}_1$) can be
  resisted (by contact forces $\bm{c}_2$ and $\bm{c}_4$), but only if a
  preload has been actively applied by the hand. In contrast, a
  disturbance pushing ``down'' ($\bm{w}_2$) will be resisted (by
  $\bm{c}_1$ and $\bm{c}_3$) passively, without the need for any
  preload, as long as the joint are not backdrivable. By correctly
  modeling energy conservation constraints, our proposed grasp model
  can capture such effects.}\label{fig:package}
\end{figure}

In robotic manipulation, passive resistance arises due to the highly
geared actuation of most commonly used hands. Large gearing ratios
between the actuators and the joints mean the fingers are
nonbackdrivable thus providing potentially rigid constraints for the
grasped object. To illustrate this concept, consider the grasp in
Fig.~\ref{fig:package}. Does the grasp remain stable if we apply
either disturbance $\bm{w}_1$ or $\bm{w}_2$ to the grasped object? In
order to resist those disturbances, contact forces must arise that
balance them. In either case, there exist contact forces that satisfy
a simple friction law (illustrated by red friction cones) and balance
the disturbance. These contact forces in turn must be balanced by
torques at the finger joints. However, it is clear that contact forces
$\bm{c}_2$ and $\bm{c}_4$ can only arise if we have previously loaded
the joints such that there is sufficient normal force at contacts 2
and 4 to sustain the friction forces required. An appropriate
\textit{preload} is required, where the actuators 'squeeze' the object
prior to the application of $\bm{w}_1$. In contrast, assuming the
joints are nonbackdrivable, contact forces $\bm{c}_1$ and $\bm{c}_3$
will arise in response to $\bm{w}_2$ irrespective of preload, and with
no action required on the part of the actuators. We refer to this
phenomenon as \textit{passive resistance}.

Passive resistance allows practitioners to greatly simplify robotic
grasping. We can often apply actuator torques that close the fingers
around the object without necessarily worrying if these will balance
out once contact is made: the fingers jam as the hand squeezes the
object, and the gearboxes between joints and actuators provide
additional structural torques. If the grasp geometry is adequately
chosen, the equilibrium joint torques arise passively when the fingers
squeeze the object between them and a stable grasp arises. For
example, in the grasp in Fig.~\ref{fig:package} it is sufficient to
actively load the joints of one finger. The nonbackdrivability of the
other finger means the object will be stably grasped and equilibrium
joint torques arise passively in the non-actuated finger.

The same phenomenon can allow a grasp to withstand a range of
disturbances applied to the object without a change in the actuator
commands. If chosen wisely, the initially applied preload is
sufficient to balance the object throughout the task and various
corresponding different disturbances. This is the true power of
passive resistance. However, while this approach works well in
practice, the field currently lacks the tools to formally analyze this
effect: we need a method that can determine which range of
disturbances will be passively resisted given a specific preload.

A grasp model that accounts for these effects must capture the
interplay of contact forces, joint torques and external wrenches. It
must be able to accurately predict how joint torques and external
wrenches are transmitted through the object and distributed across the
contacts, which is complicated by the fact that in general robotic
grasps are statically indeterminate (or
hyperstatic)~\cite{SALISBURY83}. Nonlinearities due to the unilateral
nature of contacts (a contact can only push, not pull) as well as the
nonbackdrivability of highly geared joints complicate the analysis of
such problems even in the frictionless case. As we have mentioned
earlier, the introduction of friction poses even greater difficulty,
as the Coulomb friction model is both nonsmooth and nonlinear, and
inclusion of the MDP makes it non-convex.

Despite significant advances, \remindtext{7-2-2}{no model proposed to date meets all these
requirements}. As we discuss in detail in Sec.~\ref{related_work}, some
existing models make assumptions that are not met in practical manipulation
tasks. Others approximate one or more of the constraints above, are applicable
only in two dimensions, or do not provide convergence guarantees. \addedtext{7-2-3}{Thus, the state-of-the-art grasp analysis tools cannot capture passive stability phenomena.}

In this paper, we introduce a quasi-static model that addresses all
the constraints above for general, three-dimensional grasps. \addedtext{7-2-4}{We show that inclusion of the MDP is vital to the physical accuracy and usefulness of such a model and present a convex relaxation that allows for efficient solution of problems involving the MDP.} While our
model is based on a linear approximation of friction cones, we
introduce a computationally efficient method that can successively
tighten this approximation up to the desired accuracy. \modifiedtext{7-4-1}{A very useful property of tightening
approaches is that they allow for strong guarantees}: if, at any stage of refinement,
our model fails to find a solution, we can guarantee that no solution
exists to the exact problem. This allows for early exit from
computation in cases where equilibrium cannot exist. It is, to the
best of our knowledge, the first time that a model has been proposed
that can handle three-dimensional frictional constraints that include
the MDP, up to arbitrary accuracy (and thus approaching the solution
to the exact problem) and in a computationally efficient fashion.

Depending on the choice of variables and optimization objective, our
model can be used for a wide range of queries. In this paper, we
illustrate its applicability to quasi-static grasp stability analysis
by answering multiple queries on a number of example grasps. The
queries we show here include: Given applied joint torques, will the
grasp be stable in the presence of a specified external disturbance,
assuming passive resistance effects? Alternatively, given applied
joint torques, what is the largest disturbance that can be passively
resisted in a given direction? Finally, given a disturbance, what are
the optimal joint torques that a grasp can apply for stability? We
believe these are all useful tools in the context of grasp analysis,
and plan to expand the use of this model to other types of problems in
the future.

\section{Related Work}\label{related_work}

\subsection{Grasp Force Distribution}

The most basic approaches to grasp stability analysis are concerned with the
existence of stabilizing contact forces. A grasp has \textit{force closure} if
the hand can apply arbitrary wrenches to the object through the contacts.
Analysis of force closure grasps dates back to Reuleaux
(1876)~\cite{REULEAUX1876}. Salisbury~\cite{SALISBURY_THESIS} proposed an
analytical method to test for force closure. However, perhaps the most
commonly used method to test for force closure was introduced by Ferrari and
Canny~\cite{FERRARI92} who developed an efficient geometric method for
computing the space of possible resultant wrenches a hand can impart on a
grasped object with contact forces obeying a linearized friction constraint. 
Thus, their algorithm can answer what we will call the \textit{existence
problem:} Given a disturbance applied to the grasped object, are there contact
forces that satisfy a set of simplified, linearized friction constraints, and
can balance the disturbance? In case of a positive answer their method can
furthermore determine the magnitude of the equilibrium contact forces relative
to the magnitude of a worst-case disturbance - a useful property, which has
been utilized as a grasp quality metric in numerous planning algorithms
proposed since. \addedtext{10-4-1}{Kirkpatrick et al.~\cite{KIRKPATRICK92} had
previously described a similar metric for frictionless contacts.}

However, the specific choice of grasp force, or as Bicchi~\cite{BICCHI94}
calls it the \textit{force distribution problem} is crucial for the stability
of a grasp. Salisbury et al.~\cite{SALISBURY83} investigated the conditions
for a grasp to become overconstrained for a variety of different contact types
and were the first to express contact wrenches as the sum of a particular and
a homogeneous solution. From this it followed that the static indeterminacy
could be alleviated if contact wrenches can be actively
controlled~\cite{KERR86}. This insight provided the foundation for the large
body of work on grasp force optimization, which is concerned with finding
optimal grasp forces in the space of contact forces possible under a friction
law. Given a positive answer to the \textit{existence problem} we can compute
a set of contact forces, which will balance a given disturbance and are
optimal with respect to some objective (e.g. minimizing the magnitude of the
forces.)

The first works posing the synthesis of grasp forces as a convex optimization
problem were by Kerr and Roth~\cite{KERR86} as well as Nakamura et
al.~\cite{NAKAMURA89}. While these works required a linearization of the
friction cone, later contributions~\cite{BUSS96,HAN00} have proposed
formulating the friction cone constraints as Linear Matrix Inequalities (LMI)
such that no linearization is necessary. More recently Boyd et
al.~\cite{BOYD07} formulated the problems as second-order cone problems
(SOCPs). They proposed a custom interior-point algorithm that exploits the
structure of force optimization problems and is thus very efficient.

These methods allow us to compute optimal contact forces and the joint torques
necessary to balance them. Thus, for any specific wrench on the object
encountered throughout a task we can compute the specific optimal joint
torques for stability. In order to use this in practice, however, we have to
make a string of assumptions:

\begin{itemize}

\item First, we assume perfect knowledge of the disturbance to the object
that must be balanced at all times;

\item Second, we assume that we can actively control the contact forces at
every contact;

\item Third, we assume that we can actively control the joint torques required
for equilibrium;

\item Fourth and finally, we assume that we can accurately control the torque
output of the hand actuators.

\end{itemize}

In the majority of robotic manipulation tasks these assumptions do not
hold.  First, the exact disturbance wrench acting on an object is
difficult to compute - it requires knowledge of the mass and inertial
properties as well as the exact trajectory of the object. Any
additional disturbance can not be accounted for unless the fingers are
equipped with tactile sensors. Second, many robotic hands are
kinematically deficient and contain links with limited mobility. This
means, for example, that we can not directly control the contact force
at a contact on the palm of the hand for instance. Forces at such a
contact can only arise passively by transmission of the disturbance on
the grasped object or forces at other contacts through the object.

Third, the kinematics of the hand may not permit explicit control of the
torques at every individual joint. This is the case for the class of
underactuated hands, where joint torques by definition may not be
independently controlled but are determined by the kinematic composition of
the hand. Finally, most robotic hands use highly geared motors, which makes
accurate sensing and control of the torques at the hand joints all but
impossible.

\remindtext{10-4-2}{Under these assumptions a positive answer to the
\textit{existence problem} is a necessary but by no means a sufficient
condition for the ability of a grasp to resist a given disturbance. It just
indicates the existence of contact forces that satisfy the friction laws and
can balance the applied disturbance but does not guarantee they will arise,
unless we assume that we are actively controlling the contact forces to that
end.}

A method to alleviate some of the limitations due to the above assumptions is
to take into account the flexibility of the object and the elements of the
hand. Salisbury~\cite{SALISBURY_THESIS} used this approach to derive the
stiffness matrix of a grasp and developed a framework to test for the
stability of a grasp. Nguyen~\cite{NGUYEN88} modeled each contact as a virtual
spring (first introduced by Hanafusa et al.~\cite{HANAFUSA77}) and showed that
any force closure grasp can be made stable by applying appropriate forces at
each contact. Cutkosky et al.~\cite{CUTKOSKY_COMPLIANCE} extended this work to
take into account the compliance of structural elements of the hand and object
as well as effects due to changing geometry such as contact location changes
due to rolling. However, these compliance models cannot accurately model
nonlinearities due to breaking contacts or friction.

Bicchi~\cite{BICCHI93,BICCHI94,BICCHI95} was the first to point
out the limitations of an assumption that is central to the works of early
authors such as Salisbury~\cite{SALISBURY83,SALISBURY_THESIS} and
Kerr~\cite{KERR86}. Bicchi showed that the force generation capabilities of
kinematically deficient hands are limited (no active control over forces at
palm contacts for instance.) He proposed a decomposition of the space of
possible contact forces into those actively controllable and those that may
only arise passively taking into account the kinematics of the hand. This
decomposition can be used to synthesize optimal grasp forces or to derive a
quantitative grasp quality metric. However, the compliance model
used~\cite{CUTKOSKY_COMPLIANCE} uses a linear friction model that does not
satisfy any of the constraints associated with Coulomb friction. Prattichizzo
et al.~\cite{PRATTICHIZZO97} derived grasp quality metrics using the same
compliance model but somewhat alleviated the issues of a linear friction model
by distinguishing between sticking, breaking and sliding contacts. Under their
model sliding contacts may not exhibit any frictional forces at all and hence
the results are overly conservative. Furthermore, the run-time of their 
algorithm grows exponentially with the number of contacts.

\addedtext{10-4-3}{These 'traditional' grasp stability metrics have become a
common tool in many grasp planning algorithms (to train the Dex-Net grasp
planner~\cite{MAHLER17} for instance.) Seldom, however, are actuator commands
actively controlled in response to measured disturbance as described above.
Instead, the actuators are commanded to close the fingers around an object and
often a stable grasp arises. The reasons for this lie in the design of most
robotic hands in use today: Nonbackdrivable joints provide passive resistance,
which allows for equilibrium contact forces to arise by themselves. This
phenomenon is not at all well understood.}

\subsection{Rigid Body Kinematics}

All analyses introduced thus far assume that we have some degree of control
over the contact wrenches. Palmer~\cite{PALMER_THESIS} investigated the
stability of arrangements of rigid polygonal bodies without this simplifying
assumption and showed that the problem of determining stability is co-NP
complete. Mattikalli et al.~\cite{MATTIKALLI96} pointed out that the existence
of a solution to the equilibrium equations is in fact a necessary but not
sufficient condition for stability. Pang et al.~\cite{ZAMM643} pointed out
that the equations of equilibrium are insufficient for the determination of
stability of rigid body contact problems due to the possibility of false
positives. Here lies the difficulty of the problem introduced in
Section~\ref{introduction} and Fig.~\ref{fig:package}.

Trinkle et al.~\cite{TRINKLE95} developed theory in order to predict the
motion of a rigid body in the plane under quasistatic assumptions (i.e.
inertial effects are considered negligible) and Coulomb friction -
specifically including the \textit{Maximum Dissipation Principle}
(MDP)~\cite{STEWART00}. This principle states that given relative motion at
the contact the frictional forces must do maximum work and is of foundational
importance for the modeling of rigid body mechanics. In further
work~\cite{PANG1996} Pang et al. cast the contact constraints as an uncoupled
complementarity problem (UCP) and showed that problems of this type are
NP-complete. In our own work~\cite{HAASHEGER_RSS18} we showed that exact
solutions of rigid body equilibrium with breaking and sliding contacts can be
obtained for planar grasps in polynomial time. The algorithm presented makes
use of the piece-wise convex behavior of friction forces in two dimensions and
the contact motion constraints imposed by rigid body assumptions (the latter
observation having been made by Mason~\cite{MASON_MANIPULATION}.)

For three dimensional grasps Baraff~\cite{Baraff1991,Baraff1994}
suggested iterative schemes to attempt to approach exact satisfaction
of the friction law. They however also noted that their algorithm may
not converge to find the correct solution and thus cannot provide any
guarantees. Trinkle et al.~\cite{TRINKLE97} showed that the Coulomb
friction law with maximum dissipation can be cast as a mixed nonlinear
complementarity problem (mixed NCP), which is difficult
solve. Therefore, they propose linearizing the friction cone by
approximating it as a pyramid, which allows for a formulation of the
friction constraint as a linear complementarity problem
(LCP). Problems of this type can be solved with Lemke's algorithm. A
downside of this approximation is that the friction force direction
inside the linearized cone that maximizes energy dissipation is
generally not the same one that opposes motion. Furthermore, the linearization violates the
assumption of isotropic friction. An immensely influential time
stepping scheme for the solution of multi-rigid-body dynamics with
Coulomb friction that makes use of this framework became known as the
Stewart-Trinkle formulation~\cite{STEWART96}.

Another very influential time-stepping scheme - the Anitescu-Potra
formulation~\cite{Anitescu1997} - can be obtained by omission of the
constraint stabilization term from the Stewart-Trinkle formulation. Anitescu
and Tasora~\cite{ANITESCU03,TASORA10} showed that using the LCP formulation of
the pyramidal friction cone approximation in general leads to nonconvex
solution sets. LCP problems with nonconvex solution sets contain reformulated
instances of the Knapsack problem and are therefore NP-hard. Thus, problems of
this type are difficult to solve. The authors develop an iterative algorithm
to solve them that converges to the solution of the original problem. They
achieve this through successive convex relaxation effectively solving
subproblems that have the form of strictly convex quadratic programs. They
note that their algorithm is only guaranteed to converge for 'sufficiently
small' friction coefficients but provide a lower bound for convergence. In
further work Anitescu et al.~\cite{Anitescu2010} propose a
cone-complementarity approach in order to alleviate the shortcomings of the
LCP approach with linearized cones. They develop an iterative method that
converges under fairly general conditions but may allow bodies to behave as if
they were in contact although they have drifted apart~\cite{HORAK19}. Kaufman
et al.~\cite{KAUFMANN08} showed that solving for contact forces when relative
velocities are known can be achieved by solving a quadratic program (QP).
Similarly, relative velocities can be solved for when contact forces are
known. However, problems where both are to be solved for simultaneously are
non-convex. The authors present an algorithm which iterates between the two
convex QPs until a solution of the required accuracy is found. Although it
works well in practice convergence is not guaranteed with their approach.

Todorov~\cite{TODOROV10} takes a completely new approach by deriving nonlinear
equations for the dynamic contact problem that implicitly satisfy the
complementarity conditions. In order to solve the resulting nonlinear
equations. Todorov proposes a Gauss-Newton approach with specific adaptations
such as a novel linesearch procedure. Todorov~\cite{TODOROV14} and Drumwright
et al.~\cite{Drumwright2011} independently developed formulations that are
relaxing the complementarity conditions such that a convex optimization
problem is recovered. Song et al.~\cite{SONG04} propose a compliant model of
the contacts between nominally rigid bodies. Both normal and tangential
contact forces are determined by viscoelastic constitutive relations coupling
them to local deformations. Pang et al.~\cite{PANG18} showed that the LCP formulation for discretized
friction cones first introduced by Trinkle~\cite{TRINKLE97} can be cast as a
Mixed Integer Program (MIP). While this method still suffers from the same
inaccuracies as the LCP formulation they demonstrated its sufficiency for the
control of a robotic gripper in simulation. 

In previous work~\cite{HAASHEGER_TASE18} we modeled grasps allowing for
breaking contacts using MIPs. In that work we did not treat the MDP explicitly
but instead introduced an iterative algorithm in order to mitigate
inaccuracies due to approximation of the friction law. \addedtext{7-2-5}{It
also contains a direct comparison of our method with those developed by
Ferrari and Canny~\cite{FERRARI92} as well as Prattichizzo et
al.~\cite{PRATTICHIZZO97}.} 

The work presented in this paper builds on our previous work. We demonstrate
that the MDP can be leveraged for accurate analysis of passive grasp
stability. We propose a framework for efficient solution of the resulting
problems by approximating the friction law including the MDP and successively
refining the approximation to obtain solutions arbitrarily close to the exact
solution.

\section{Grasp Model}\label{model}

Consider a robotic hand that makes $m$ contacts with a grasped
object. The systems is initially at rest and we would like to
determine if the system will remain at rest when a disturbance is
applied to the object. Each contact is defined by a location on the
surface of the grasped object and a normal direction determined by the
local geometry of the bodies in contact. We chose the \textit{Point
  Contact with Friction} model to describe the possible contact
wrenches that may arise at the interfaces between the hand and the
object. Therefore we only consider contact forces and do not allow for
frictional torques. This is a reasonable assumption for contacts
between smooth and relatively stiff bodies. For any contact specific
vector (such as the contact force), we will use subscripts $n$ and $t$
respectively to denote the components lying in the contact normal and
contact tangent directions. We use the vector $\bm{c} \in
\mathbb{R}^{3m}$ to denote contact forces, where $\bm{c}_i \in
\mathbb{R}^3$ is the force at the $i$-th contact. Using the notation
above, $c_{i,n} \in \mathbb{R}$ is the normal component of this force,
and $\bm{c}_{i,t} \in \mathbb{R}^2$ is its tangential
(i.e. frictional) component.

\vspace{2mm} \noindent \textbf{Equilibrium:} The grasp map matrix $\bm{G} \in
\mathbb{R}^{6 \times 3m}$ maps contact wrenches into a frame fixed to the
grasped object. We can now write the equilibrium equations for the grasped
object where we collect all disturbances externally applied to the object
(such as gravitational forces for instance) as $\bm{w} \in \mathbb{R}^6$. For
a hand with $l$ joints the transpose of the grasp Jacobian $\bm{J} \in
\mathbb{R}^{3m \times l}$ relates contact forces to the torques $\bm{\tau} \in \mathbb{R}^l$ in
the hand joints required for hand equilibrium.
\begin{alignat}{2}
\bm{G}\bm{c} &+ \bm{w} &~= 0 \label{eq:wrench_eq}\\
\bm{J}^T \bm{c} &+ \bm{\tau} &~= 0 \label{eq:torque_eq}
\end{alignat} 

These equilibrium equations can predict the resultant wrench on the object and
the joint torques required to balance given contact forces, but, since neither
$\bm{G}$ nor $\bm{J}^T$ are typically invertible, they have no predictive
capabilities in the opposite directions: we can not use them to predict the
contact response to known external wrenches, or joint torques. They thus fail
to capture effects where contact forces are transmitted (either amongst the
joints or between the joints and the external environment) through the object
itself, and cannot account for many of the phenomena illustrated in
Section~\ref{introduction}. 


\vspace{2mm} \noindent \textbf{Virtual motion:} To resolve the
structural indeterminacy of the grasp and determine which contact
forces will arise in response to a disturbance, we must introduce
additional constitutive relations. Following the grasp compliance
model introduced by Cutkosky and Kao~\cite{CUTKOSKY_COMPLIANCE} and
later used by Bicchi~\cite{BICCHI93,BICCHI94,BICCHI95}, we model contact
stiffness by introducing virtual springs at contacts. Normal forces
must arise as a result of these springs being loaded through
\textit{virtual motion} of the grasped object. However, \addedtext{2-2}{we} significantly extend these
models, by including constraints for
unilateral contacts, and a more accurate friction model.

The constitutive relations we use relate contact forces to the
relative \textit{virtual motion between the hand and the object at the
  contacts}. This relative motion, expressed in the contact frames, is
denoted by $\bm{d} \in \mathbb{R}^{3m}$ (only the translational
components are of interest to us), and can be formulated in terms of
the overall virtual object motion $\bm{r} \in \mathbb{R}^6$ and the
virtual joint motion $\bm{q} \in \mathbb{R}^l$:
\begin{equation}
\bm{G}^T\bm{r} - \bm{J}\bm{q} = \bm{d} \label{eq:rel_contact_motion}
\end{equation}

The introduction of motion in our framework might initially seem to
contradict our focus on the
static equilibrium of a grasp. However, this allows us to resolve the
structural indeterminacy of the grasp through the use of constitutive
relations. It furthermore allows us to enforce nonsmooth constraints
such as the unilaterality of the contacts and the non-backdrivability
of the joints. One can think of our approach as solving the first step
of the dynamics of the hand-object systems in order to determine if it
is in equilibrium. Thus, we consider all motion to be 'virtual', used
as a tool to enforce results that are consistent with rigid body
behaviors: \textit{all contact forces must be consistent with some
  virtual motion of the grasped object and the joints, which become
  additional variables in our framework.}


\vspace{2mm} \noindent \textbf{Normal forces:} The constitutive
relation for normal forces assumes virtual springs of
stiffness $k$ along the contact normals. Thus, the normal force at a
contact $i$ is determined by the relative motion between the object
and the robot hand at that contact in the direction of the contact
normal. However, we also model unilateral contacts, which may only
push on an object but can never pull (we do not concern ourselves with
adhesion or similar effects).  Thus, the normal force at a contact
must be strictly non-negative.  Furthermore, if the contact detaches
the contact force must be zero. For simplicity and without loss of
generality we can assume $k=1$.
\begin{equation}
\begin{cases}
c_{i,n} = -d_{i,n} & \text{if } d_{i,n} \leq 0 \\
c_{i,n} = 0 & \text{if } d_{i,n} > 0
\end{cases}\label{eq:normal_forces}
\end{equation}

\begin{figure}[t!]
\centerline{
\subfloat[Coulomb friction cone]{\includegraphics[width=0.46\columnwidth]{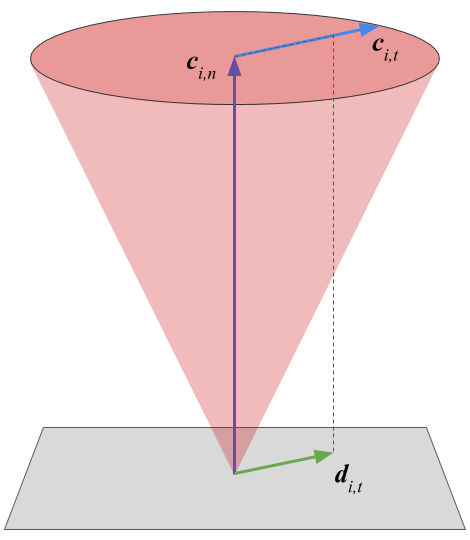}\label{fig:pyramid_a}}\hfill%
\subfloat[Pyramidal approximation]{\includegraphics[width=0.46\columnwidth]{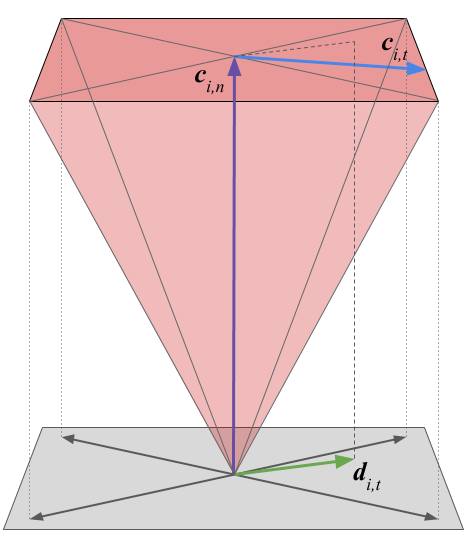}\label{fig:pyramid_b}}}
\caption{\addedtext{10-1-1}{Illustration of the exact Coulomb friction cone and a pyramidal approximation as we use it. The black arrows positively span the contact tangent plane and make up the matrix of basis vectors $\bm{D}_i$.}}\label{fig:pyramid}
\end{figure}

\vspace{2mm} \noindent \textbf{Friction forces:} To model friction, we chose
the Coulomb model. The first part of the Coulomb model provides an upper bound
to the magnitude of the friction force given the normal force and the friction
coefficient $\mu_i$ (as all motion is virtual we consider all friction
coefficients to be those of static friction.) This defines a cone
$\mathcal{F}_i$ at each contact (see Fig.~\ref{fig:pyramid_a}.)
\begin{equation}
\mathcal{F}_i( \mu_i, c_{i,n} ) 
= \{ \bm{c}_{i,t} : \left\lVert \bm{c}_{i,t} \right\rVert \leq \mu_i c_{i,n} \}
\label{eq:friction_bound}
\end{equation}

The second part of the Coulomb model concerns the Maximum Dissipation
Principle (MDP)~\cite{STEWART00}: Given a relative contact motion, the
friction force at that contact must maximize energy dissipation, while
bounded by eq. (\ref{eq:friction_bound}).
\begin{equation}
\bm{c}_{i,t} \in \argmin_{\bm{c}_{i,t} \in \mathcal{F}_i} \bm{c}^T_{i,t} \cdot \bm{d}_{i,t}
\label{eq:mdp}
\end{equation}
In the case of isotropic friction, the dissipation is maximized if the friction
force is anti-parallel to the relative sliding motion and lies on the boundary
of the cone $\mathcal{F}_i$. Thus, we can also directly express the friction
force in terms of the normal force and the relative sliding motion. We must
distinguish between two cases: 

\begin{itemize}
\item At a contact that does not exhibit
relative motion in a tangential direction (sliding) the friction force is
constrained such that the contact force lies within the cone $\mathcal{F}_i$
\item If a contact does exhibit sliding, the friction force must oppose the direction of motion (or incipient acceleration), and the total contact force must lie on the friction cone edge. 
\end{itemize}
The complete model can be formulated as follows:
\begin{equation}
\begin{cases}
\left\lVert \bm{c}_{i,t} \right\rVert \leq \mu_i c_{i,n}, & \text{if}~\left\lVert \bm{d}_{i,t} \right\rVert = 0\\
\bm{c}_{i,t} = - \mu_i c_{i,n} \frac{\bm{d}_{i,t}}{\left\lVert\bm{d}_{i,t}\right\rVert}, & \text{otherwise}
\end{cases}\label{eq:fric_forces}
\end{equation}

Note that the formulation in eq. (\ref{eq:fric_forces}) requires the
distinction between sliding contacts and those remaining at rest while the
original formulation in eqs. (\ref{eq:friction_bound})\&(\ref{eq:mdp}) holds
in both cases. For isotropic friction they are equivalent, and we
have thus arrived at an accurate constitutive relation describing the friction
forces in terms of the relative contact motion. 

\vspace{2mm} \noindent \textbf{Joint torques:} Finally, we model the joints as
non-backdrivable, in order to capture the behavior of the majority robotic
hands driven by highly geared motors. This means that a joint $j$ may only
exhibit virtual motion in the direction that its commanded torque $\tau_{j,c}$
is moving it in. The joint torque may exceed the commanded level, but only if
this arises passively. This means that a joint that is being passively loaded
beyond the commanded torque levels must be locked in place and may not move. A
moving joint must apply the torque it was commanded to. Thus, we must also
distinguish between two states for each joint.
\begin{equation} 
q_j \geq 0,~ \begin{cases} 
\tau_j \geq \tau_{j,c},& \text{if}~q_j = 0\\
\tau_j = \tau_{j,c}, & \text{if}~q_j > 0
\end{cases}\label{eq:joint_model}
\end{equation}
A joint with zero commanded torque may not move, as any torque arising
from external factors will be absorbed by the gearing.

\vspace{2mm} \noindent \textbf{Complete problem:} The system comprising eqs.
(\ref{eq:wrench_eq})-(\ref{eq:joint_model}) defines the static equilibrium
formulation for a grasp. It is very general in nature, and can be considered
as part of existence problems (e.g. given $\bm{\tau}$, determine if $\bm{r}$
and $\bm{c}$ exist that balance a given $\bm{w}$), or optimization problems,
with the addition of an objective (e.g. determine the optimal $\bm{\tau}$ that
satisfies the existence problem above). Remaining agnostic to the exact query
that is being solved we will refer to the \textit{exact problem} as the
following query: given a subset of $\bm{\tau}$, $\bm{r}$, $\bm{c}$ or
$\bm{w}$, determine the rest of these variables such that
(\ref{eq:wrench_eq})-(\ref{eq:joint_model}) are exactly satisfied.

The main difficulty of directly solving this exact problem lies in the
non-convexity of the friction law when including the MDP \addedtext{7-4-2}{-
the constraints in (\ref{eq:fric_forces}) contain a nonlinear equality}. Many of
the works we have reviewed earlier seek to find approximations that lend
themselves to efficient solvers. The compliance grasp
model~\cite{CUTKOSKY_COMPLIANCE} makes a linear approximation, while
Prattichizzo et al.~\cite{PRATTICHIZZO97} also make a distinction between
sticking and sliding contacts. However, due to the non-convexity of the second
part of (\ref{eq:fric_forces}) they must make the overly conservative
approximation that a sliding contact may not apply any frictional forces at
all. 

\addedtext{7-4-3}{Of course one could cast the MDP in its original
  minimization formulation instead: If we remove
  (\ref{eq:fric_forces}) from the exact problem and introduce instead
  a minimization objective (\ref{eq:mdp}) then an optimal solution to
  this new problem would be valid if it also satisfies
  (\ref{eq:fric_forces}). However, this formulation is of course still
  non-convex as the bilinear form in (\ref{eq:mdp}) is not positive
  definite~\cite{LIBERTI04}. However, this reformulation of the exact
  problem would allow for a global optimization approach based on a
  convex relaxation of the bilinear forms using McCormick
  underestimators~\cite{McCormick1976}, which relies on upper and
  lower bounds on the variables involved in the bilinear forms. We
  could then use a Spacial Branch-and-Bound (sBB)
  algorithm~\cite{TUY16} to solve for the global optimum (see for
  example the $\alpha$BB algorithm~\cite{FLOUDAS98} or the
  BARON~\cite{BARON} software for global optimization.)}

\addedtext{7-4-4}{In this paper we make use of the particular
  structure of the grasping problem to derive a relaxation that does
  not require variable bounds, since in the grasping problem bounds on
  contact forces and particularly virtual motions are not implicit to
  the problem and thus difficult to predetermine. Furthermore, we relax
  the constraints of the optimization problem instead of the
  objective. This allows us to use the objective to solve for
  interesting grasp characteristics such as the optimum actuator
  commands.}

The friction models used in the grasp force optimization
literature~\cite{SALISBURY83,KERR86,NAKAMURA89,BUSS96,HAN00,BOYD07} assume
full control over the contact forces and therefore only have to concern
themselves with stationary contacts. This removes the non-convexity that
arises from the MDP. As all motion in our problems is virtual, one may indeed
be tempted to ignore the sliding effects and use one of the friction models in
the grasp force optimization literature. We can achieve this by solving the
problem described by equations (\ref{eq:wrench_eq})-(\ref{eq:friction_bound}),
(\ref{eq:joint_model}) and neglecting equation (\ref{eq:mdp}). This approach,
however, fails to capture the passive effects in the grasps we want to
analyze. As we will show in Section~\ref{results} solving
(\ref{eq:wrench_eq})-(\ref{eq:friction_bound}), (\ref{eq:joint_model}) without
the MPD results in unphysical solutions that violate the laws of energy
conservation. In contrast, as we will also show in Section~\ref{results},
including the MDP in our formulation allows us to obtain physically meaningful
results.

The approaches from the rigid body dynamics literature are perhaps more
applicable to these problems, as they concern themselves with moving bodies
and therefore must include some treatment of the MDP. However, due to the
difficulty in solving the resulting
problems~\cite{PALMER_THESIS,ANITESCU03,KAUFMANN08} they are either
computationally infeasible~\cite{TRINKLE95,PANG1996} or \remindtext{10-5}{make
approximating
assumptions~\cite{TRINKLE97,STEWART96,Anitescu1997,Anitescu2010,TODOROV10,TODOROV14,Drumwright2011,SONG04,PANG18}
in order to allow for real-time simulation}. Others make no convergence
guarantees~\cite{Baraff1991,Baraff1994,HAASHEGER_TASE18} or are only
guaranteed to converge for small friction coefficients~\cite{TASORA10}.

\textit{To the best of our knowledge, no efficient solution has been proposed
to date for a three-dimensional grasp model that includes an exact formulation
of Coulomb friction such as eq. (\ref{eq:fric_forces}), or an equivalent
reformulation. Our main contribution is a method to efficiently find solutions
to progressively tighter approximations of this model, up to arbitrary
accuracy.}

\begin{figure*}[t!]
\centerline{\subfloat[Exact]{\includegraphics[width=1.12in]{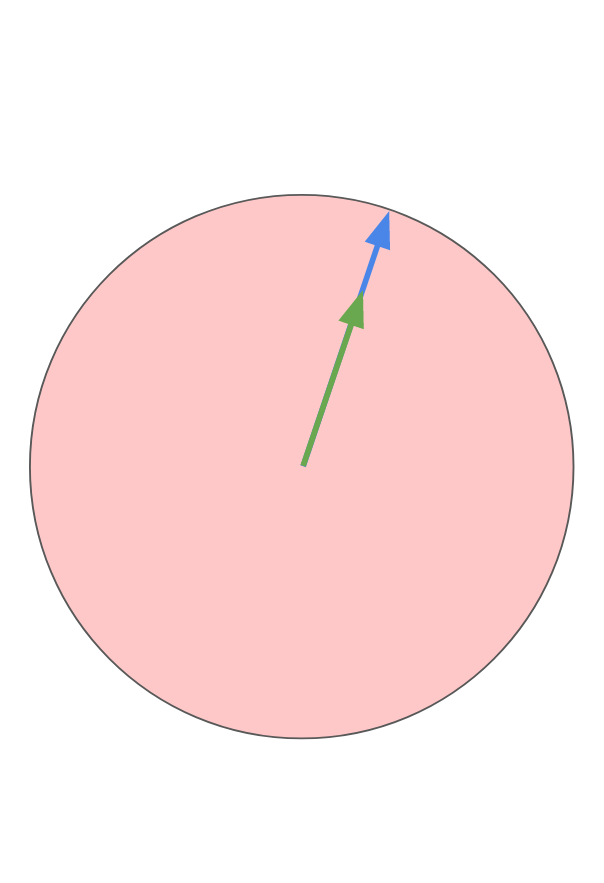}
\label{fig:cone4}}
\qquad
\subfloat[Step 1]{\includegraphics[width=1.75in]{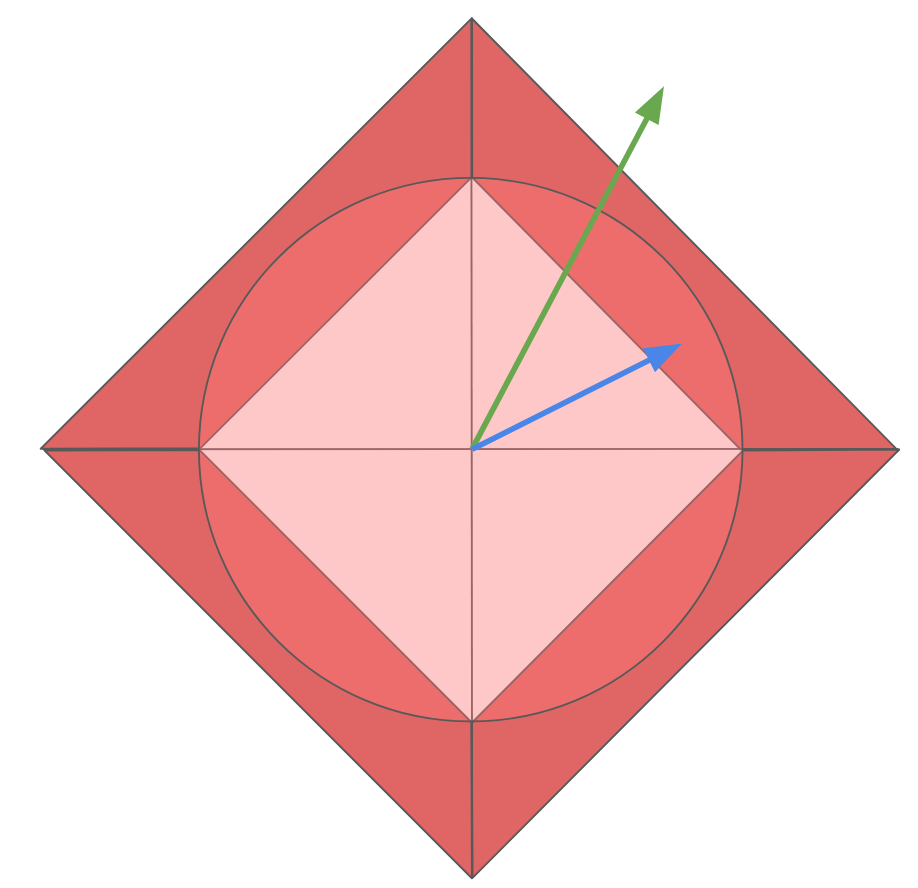}%
\label{fig:cone1}}
\qquad
\subfloat[Step 2]{\includegraphics[width=1.75in]{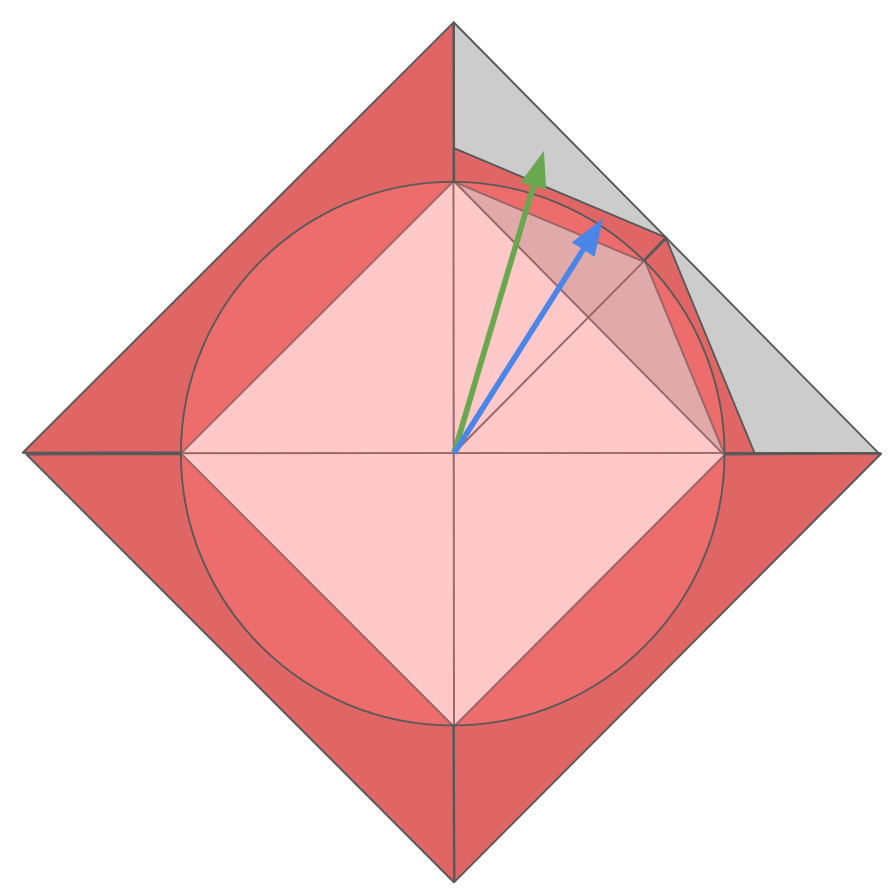}%
\label{fig:cone2}}
\qquad
\subfloat[Step 3]{\includegraphics[width=1.75in]{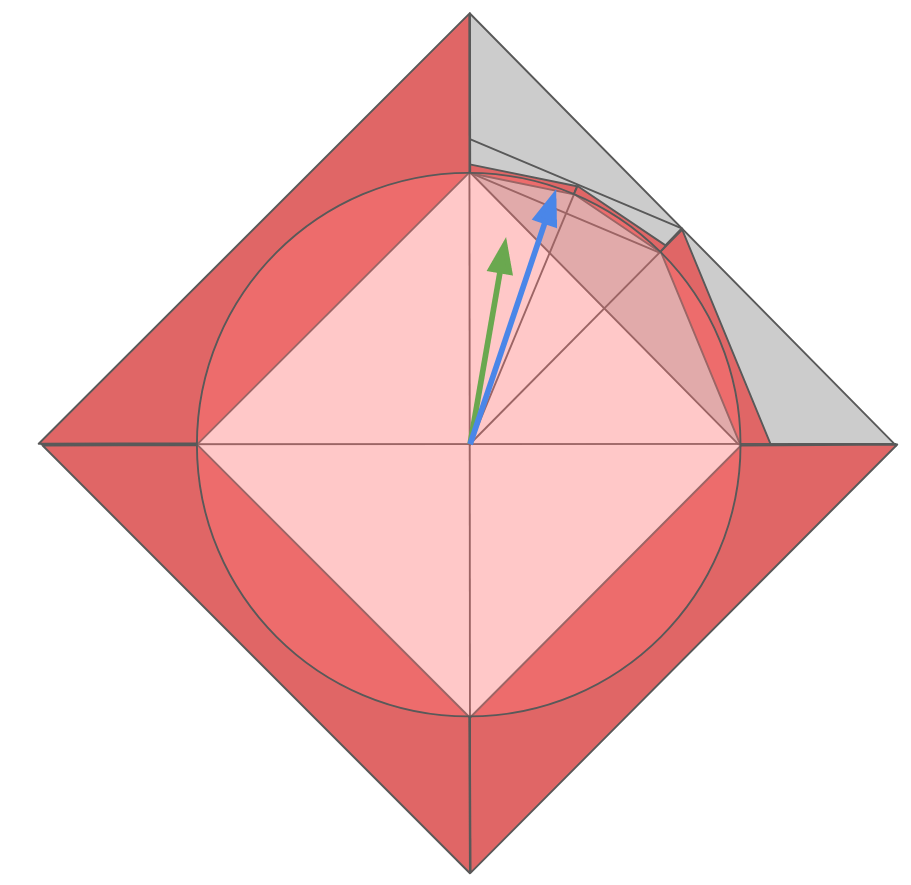}%
\label{fig:cone3}}}
\caption{First three steps of our algorithm refining the friction law relaxation successively and locally. The dark red regions are the space of feasible friction forces at a sliding contact. If there is no tangential motion the light red regions are added to the feasible friction force space. The green and blue arrows are $\bm{d}_{i,t} = \bm{D}_i \bm{\alpha}_i$ and $\bm{c}_{i,t} = \bm{D}_i \bm{\beta}_i$ respectively, which drive refinement of their local friction cone sector.}\label{fig:cone}
\end{figure*}

\section{Formulation as Mixed Integer Program}\label{solution}

Let us take a closer look at the grasp model constraints introduced so far.
Constraints (\ref{eq:wrench_eq})-(\ref{eq:rel_contact_motion}) are simple
equality constraints. We note that the constraints in (\ref{eq:normal_forces})
and (\ref{eq:joint_model}) exhibit a combinatorial nature - a type of
constraint that can be found in mixed integer programming. In fact both the
normal force and the joint model relationships can be cast as pairs of convex
constraints with binary decision variables in an MIP that can then be solved
using algorithms such as branch and bound. The friction law is more complex:
the constraint for sliding contacts in (\ref{eq:fric_forces}) defines a
non-convex set. Therefore, we must formulate a relaxation of this constraint
such that we can solve the system as an MIP.

We start from the common linearized friction model which replaces the circular
friction cone at contact $i$ with its discretization  as a polygonal cone (see
Fig.~\ref{fig:pyramid}.) Matrix $\bm{D}_i \in \mathbb{R}^{3 \times k}$
contains as its columns a set of $k$ vectors that positively span the contact
tangential plane and thus the space of possible friction forces. Frictional
forces can now be expressed as a positive linear combinations of these so
called \textit{friction edges} with weights $\bm{\beta}_i \in \mathbb{R}^k$.
Inequality constraints on a vector are to be understood in a piecewise
fashion.
\begin{equation}
\bm{D}_i \bm{\beta}_i = \bm{c}_{i,t},~\bm{\beta}_i \geq 0\label{eq:fric_discretization}
\end{equation}
We can express relative tangential contact motion as a weighted combination of
the same friction edges, with weights $\bm{\alpha}_i$. For reasons that will
soon become apparent, we choose to express the negative of the tangential
motion instead:
\begin{equation}
\bm{D}_i\bm{\alpha}_i = -\bm{d}_{i,t},~\bm{\alpha}_i \geq 0\label{eq:alpha_movement}\\
\end{equation}

\addedtext{10-1-3}{If the friction edges in $\bm{D}_i$ are arranged in an
ordered fashion such that neighboring friction edges in the tangent plane are
also neighbors in $\bm{D_i}$} we can constrain friction to (approximately)
oppose motion by requiring that \textit{the friction force lies in the same
sector of the linearized friction cone as the negative of the tangential
contact motion.} Without loss of the above properties, we require that at most
two components of $\bm{\beta}_i$ can be non-zero and that non-zero components
are either consecutive or lie at the first and last positions of vector
$\bm{\beta}_i$. This can be achieved by constraining $\bm{\beta}_i$ with a
special ordered set $\bm{z}_i \in \mathbb{R}_{\geq 0}^{k+1}$ of type 2
(SOS2)~\cite{Beale1976}, which has one more component than $\bm{\beta}_i$
itself. 
\begin{equation}
\beta_{i,1} \leq z_{i,1} + z_{i,k+1},~\beta_{i,2} \leq
z_{i,2},~...,~\beta_{i,k} \leq
z_{i,k},~\bm{z}\in\text{SOS2}\label{eq:sos_beta} \end{equation}
\addedtext{10-1-2}{A special ordered set of type 2 is a set of ordered
non-negative numbers of which at most two can be non-zero. If two numbers in
the SOS2 are non-negative they must be consecutive in their ordering.} This
type of constraint can be solved by MIP solvers and is hence admissible to our
model. We now similarly constrain the weights $\bm{\alpha}_i$ that determine
relative motion with the same SOS2 as in (\ref{eq:sos_beta}).
\begin{eqnarray}
\alpha_{i,1} \leq z_{i,1} + z_{i,k+1},~\alpha_{i,2} \leq z_{i,2},~...,~\alpha_{i,k} \leq z_{i,k}\label{eq:sos_alpha}
\end{eqnarray}

\addedtext{10-1-4}{Note that these constraints hold for both sliding as well
as stationary contacts. If a contact is stationary then
all components of $\bm{\alpha}_i$ must be zero. Thus, any two consecutive
components of $\bm{z}_i$ may be non-zero and since $\bm{\beta}_i$ are the
weights of the basis vectors in $\bm{D}_i$ the friction force may point in any
direction in the tangent plane. If the contact slides then some components of
$\bm{\alpha}_i$ must be nonzero. As both $\bm{\alpha}_i$ and $\bm{\beta}_i$
are constrained by $\bm{z}_i$ and only two consecutive components of
$\bm{z}_i$ may be nonzero, this means the friction force must lie in the same
sector of the friction pyramid as the negative of the tangential relative
contact motion, but may not be collinear (see Fig.~\ref{fig:pyramid_b}.)}

Finally, we can constrain the magnitude of the friction force in addition to
its direction. For sliding contacts, friction must be maximized, while for
stationary contacts it only has an upper bound. The friction edges in
$\bm{D}_i$ are chosen to be unit vectors such that these constraints can be
expressed as follows:
\begin{equation}
\begin{cases}
\bm{e}^T \bm{\beta}_i \leq \mu_i c_{i,n}, & \text{if}~\left\lVert \bm{d}_{i,t} \right\rVert = 0\\
\bm{e}^T \bm{\beta}_i = \mu^i c_{i,n}, & \text{otherwise}
\end{cases}\label{eq:fric_approx}
\end{equation}
where $\bm{e} = (1, 1, ..., 1) \in \mathbb{R}^k $. Constraint
(\ref{eq:fric_approx}) can also be included in an MIP using a binary decision
variable.

We now have a complete model of friction. For a finite value of $k$, this
model is approximate; in the limit, as $k \rightarrow \infty$ eq.
(\ref{eq:fric_approx}) in combination with constraints (\ref{eq:sos_beta} -
\ref{eq:sos_alpha}) behave like the  Coulomb friction model in
(\ref{eq:fric_forces}).

\section{Successive Hierarchical Refinement}\label{refinement}

We can solve the complete system described by constraints
(\ref{eq:wrench_eq})-(\ref{eq:normal_forces}),
(\ref{eq:joint_model})-(\ref{eq:fric_approx}) as an MIP with algorithms such
as branch and bound. In order to improve our approximation, we could choose a
high number of edges for the discretized friction cones. In practice, however,
that approach is not feasible as the time taken to solve an MIP is sensitive
to the number of integer variables in the problem.  As SOS2 constraints are
implemented using binary variables a highly refined friction cone
approximation quickly becomes computationally intractable.

\subsection{Intuition}

Our approach is based on the key insight that one can obtain an equally accurate solution by solving a problem
with a coarse friction cone approximation, and \textit{successively refining the
linearized friction constraints only in the region where friction
forces arise.}  Our approach thus proceeds as follows:
\begin{itemize}
\item We solve our problem using a coarse approximation of the friction cone (few friction edges). 
From the solution, we identify the sector of the linearized cone (the area between two edges) where 
both the friction force and the negative of the relative motion lie.
\item To obtain a tighter bound, we add new friction edges that refine \textit{only the sector identified above}. 
We then repeat the procedure with the new, selectively refined version of the friction cone.
\end{itemize}

An important characteristic of our method is that we can choose our friction edges so that, at any level of refinement, \textit{the solution set to the approximate problem contains the solution to the exact problem}, assuming one exists. \addedtext{7-4-6}{This concept is a necessary condition for a tightening approach} and is visually illustrated in Fig.~\ref{fig:cone}. For the exact problem, the feasible space of friction forces for sticking contacts is the inside of the circle, while the feasible space of friction forces for sliding contacts is the circle itself - a non-convex set. Consider now a rough approximation with four friction edges (Step 1). If we allow sticking friction to reside inside the areas shaded in either shade of red, while sliding friction must lie within the dark red border, the space of allowable solutions to the exact problem is contained inside our linear and piecewise convex approximation. Assume that, at this level of refinement, there is a solution to an equilibrium problem, with friction force lying inside the upper right sector. We refine this sector, again taking care that the space of allowable solutions to the exact problem is contained inside our refinement (Step 2). We continue this procedure (Step 3, etc.) until one of two things happen: we either reach a level of refinement where no solution exists, or we refine down to the point where the active sector is as small as we want it to be, bringing us arbitrarily close to the solution to the exact problem.

This refinement scheme provides two important advantages: If, at any point during the refinement, no solution exists that satisfies equilibrium, we can guarantee that no solution can exist to the exact version of the problem either. This guarantee immediately follows from the properties that \remindtext{2-4-1}{the solution set at any refinement level includes the solution set at the next level}, and that, in the limit, our discretization approaches the exact constraints. (Note that alternative discretizations of the friction constraints, such as the LCP formulations discussed previously, do not exhibit this property.) In practice, this means that, when no solution exists to the exact equilibrium problem, our algorithm can determine that very quickly, only solving relatively coarse refinement levels.

The second advantage our scheme provides is that when a solution does exist, we can typically refine it to high accuracy (a very close approximation to the solution of the exact problem) using relatively few friction edges. This is not theoretically guaranteed: in the worst case, our approach could require all sectors to be fully refined before finding an adequate solution as the desired resolution, and may hence perform worse than using a fully refined friction discretization to begin with. However, we have never found that to be the case. Typically, only a small region of the discretization must be refined as the contact forces are also constrained by equilibrium relations (\ref{eq:wrench_eq}) \& (\ref{eq:torque_eq}) and will generally point in similar directions at all levels of refinement, leading to a very localized and targeted tightening of the relaxation. Thus, this algorithm is efficient enough to analyze complex grasps on a consumer PC to levels of refinement that are otherwise unachievable.

However, in order to achieve these abilities, we have to implement a friction cone refinement method meeting the requirement discussed above. We present the details of our chosen refinement scheme next.

\subsection{Implementation}

\begin{algorithm}[t]
\caption{Grasp analysis through successive relaxation}\label{alg:refinement}
\begin{algorithmic}[0]
\Procedure{Relaxation refinement}{}
  \State \textbf{Input:}
  \State ~~~~$O$ - objective function
  \State ~~~~$C$ - additional constraints
  \State ~~~~$\gamma$ - initial refinement level (angle btw. friction edges)
  \State ~~~~$q$ - max refinement level desired
  \State Initialize $\bm{D}$ with basis vectors of length $l_1$ as in (\ref{eq:l1})
  \Do
    \State Optimize $O$ subject to (\ref{eq:wrench_eq})-(\ref{eq:normal_forces}), (\ref{eq:joint_model})-(\ref{eq:sos_alpha}), (\ref{eq:fric_refine}) and $C$.
    \If{no solution exists}
      \State \textbf{return} no feasible solution
    \EndIf
    \State $\text{refinement\_needed} \gets \texttt{False}$
    \For{each contact $i$}
      \State Find active edges in $\bm{D}_i$
      \State $\delta_i \gets \text{angle between active edges}$
      \If{$\delta_i > \gamma / 2^q$}
        \State $p \gets \log_2 (\gamma / \delta_i) + 2$
        \State Add new edges to $\bm{D}_i$ of length $l_p$ as in (\ref{eq:length})
        \State Remove redundant edges from $\bm{D}_i$
        \State $\text{refinement\_needed} \gets \texttt{True}$
      \EndIf
    \EndFor
  \doWhile{$\text{refinement\_needed}$}
\State \textbf{return} solution
\EndProcedure
\end{algorithmic}
\end{algorithm}

Let us pick the initial basis vectors in $\bm{D}$ such that the angle $\gamma$
between all pairs of successive vectors is equal. We pick an initial angle
$\gamma = \pi / 2$. We refine our polyhedral friction cones by bisecting
sectors defined by the non-zero components of $\bm{z}$ and define the
angle at which to stop refinement as $\gamma / 2^q$. We now find the required
length $l_1$ of these initial friction edges such that the initial solution
set contains the solution sets at all refinement levels.
\begin{equation}
l_1 = \prod_{r=1}^{q+1} \sec(\frac{\gamma}{2^r})
\label{eq:l1}
\end{equation}

Thus, we modify the friction edges in $\bm{D}$ accordingly. We define vector $\bm{f}_i$ to
contain the lengths of the friction edges making up the friction cone
approximation at a contact in an order corresponding to the order of the
weights $\bm{\beta}_i$. Constraints (\ref{eq:fric_approx}) now become
\begin{equation}
\begin{cases}
\bm{e}^T \bm{\beta}_i \leq \mu_i c_{i,n}, & \text{if}~\left\lVert \bm{d}_{i,t} \right\rVert = 0\\
\bm{e}^T \bm{\beta}_i \leq \mu_i c_{i,n},~\bm{f}_i^T \bm{\beta}_i \geq \mu_i c_{i,n}, & \text{otherwise}
\end{cases}\label{eq:fric_refine}
\end{equation}

We are now ready to solve the initial coarse relaxation problem defined by
(\ref{eq:wrench_eq})-(\ref{eq:normal_forces}), (\ref{eq:joint_model})-(\ref{eq:sos_alpha}) and
(\ref{eq:fric_refine}). We find the two active friction edges $d_1$ and $d_2$ and create three new edges that point in the direction of $d_1$, $d_1+d_2$ and $d_2$ and have magnitude $l_2$. For this and all following refinements we have 
\begin{equation}
l_p = \prod_{r=p}^{q+1} \sec(\frac{\gamma}{2^r})\label{eq:length}
\end{equation}
where $p$ is the level of refinement of the sectors to be created. We insert
the new friction edges between $d_1$ and $d_2$ in matrix $\bm{D}_i$ and remove
any redundant friction edges (edges that are identical or edges that lie
between any such edges). Solving the problem thus obtained we can continue
refining the friction discretization until the angle of the active sectors at
all contacts reach an angle of $\gamma / 2^q$. We do not further refine any
sector that has already reached this threshold. The overall method is shown in
Algorithm~\ref{alg:refinement}.

\addedtext{7-4-5}{Note that this convex relaxation does not rely on explicit
bounds on the problem variables as would be the case were we using McCormick
envelopes~\cite{McCormick1976}. Instead, we use the bounds implicit in the
friction cone constraint (\ref{eq:fric_forces}) in our relaxation. This is also a further advantage
of our approach as the relaxation has an easily understood physical
interpretation.}

\section{Improving Robustness to Geometrical Uncertainties}\label{uncertainties}

In the above we outlined a grasp model that allows us to analyze the grasp
stability given perfect information about the geometry of the grasp. We assume
we know exactly the contact position and orientation. In practice however we
often encounter uncertainties, which can greatly affect the stability of a
grasp. Even when using tactile sensors in order to locate contacts made
between the hand and the object the contact normals (and hence orientation)
are often difficult to obtain accurately. Therefore we would like to make our
framework robust to discrepancies up to a certain magnitude. We introduce the method 
we use for this here, and illustrate its importance in the following section.

Let us suppose we have an upper bound on our uncertainty in the contact normal
$\eta$. Thus, the actual contact normal lies in a space of possible contact
normals that deviate by at most angle $\eta$ from the nominal contact normal.
In order for a grasp to be robust to deviations defined by this space we would
like it to be robust in the worst-case. The worst-case contact normal is one
such that its projection into the tangent plane opposes the relative
tangential contact motion. In our space of contact normals such a normal would
be the most effective at unloading the contact and hence destabilizing the
grasp. The relative contact motion in direction of the worst-case normal
would then be given by
\begin{equation}
\hat{d}_{i,n} = {d}_{i,n} \cos(\eta) - \left\lVert \bm{d}_{i,t} \right\rVert \sin(\eta)
\label{eq:uncertainty}
\end{equation}

However, we need to find a linear approximation for $\left\lVert \bm{d}_{i,t}^T
\right\rVert$ as including it exactly would introduce a nonconvex quadratic
equality constraint. Fortunately we can use the amplitudes of the friction
edges $\alpha$. Using (\ref{eq:alpha_movement}) \& (\ref{eq:length}) the
summation of the product of all contact motion amplitudes with the length of
the corresponding friction edges gives us an estimate of the magnitude of the
relative tangential contact motion.
\begin{equation}
\left\lVert \bm{d}_{i,t} \right\rVert \approx \sum_{s=1}^{k} l_p \alpha_{i,s}\label{eq:overestimate}
\end{equation}
where $p$ is the level of refinement of the active sector (i.e. the sector corresponding to the
nonzero components of $\bm{\alpha}_i$.) This formulation is equivalent to one
where all the friction edge vectors in $D$ are of unit length where we could
then omit $l_p$. 

The problem with this formulation is that, at low resolutions of the friction
cone, it overestimates the relative tangential contact motion. It is exact for
contact motion parallel to to any friction edge (tangential motion of
magnitude 1 will result in exactly one component of $\bm{\alpha}$  equal to
1). However, for any unit tangential motion that lies between two edges, the
sum of the two active components of $\alpha$ must be greater than 1 due to the
triangle inequality. This effect diminishes at finer resolutions as friction
edges become closer to parallel, but the destabilizing effect is potentially
larger at coarser resolution.

However, recall that our refinement method requires that the solution set at 
coarser levels includes the solution set at more detailed levels. We thus require 
the destabilizing effect to be weaker at coarse resolutions and become stronger
approaching its exact value as $k \rightarrow \infty$. Therefore we
modify (\ref{eq:overestimate}) such that it underestimates tangential motion,
except at the midpoint between two edges where it is exact.
\begin{equation} 
\left\lVert \bm{d}_{i,t} \right\rVert \approx
\frac{l_p}{l_{p+1}} \sum_{s=1}^{k} l_p \alpha_{i,s} 
\end{equation}
As $k \rightarrow \infty$ this estimation becomes exact. We now replace the
normal relative contact motion in (\ref{eq:normal_forces}) with
$\hat{d}_{i,n}$ in order to obtain solutions that are robust to uncertainties
in contact normal up to an angular discrepancy of $\eta$. 
\begin{figure}[tbp]
\centering
   \begin{subfloat}[View from the side]
   {\includegraphics[width=1.0\columnwidth]{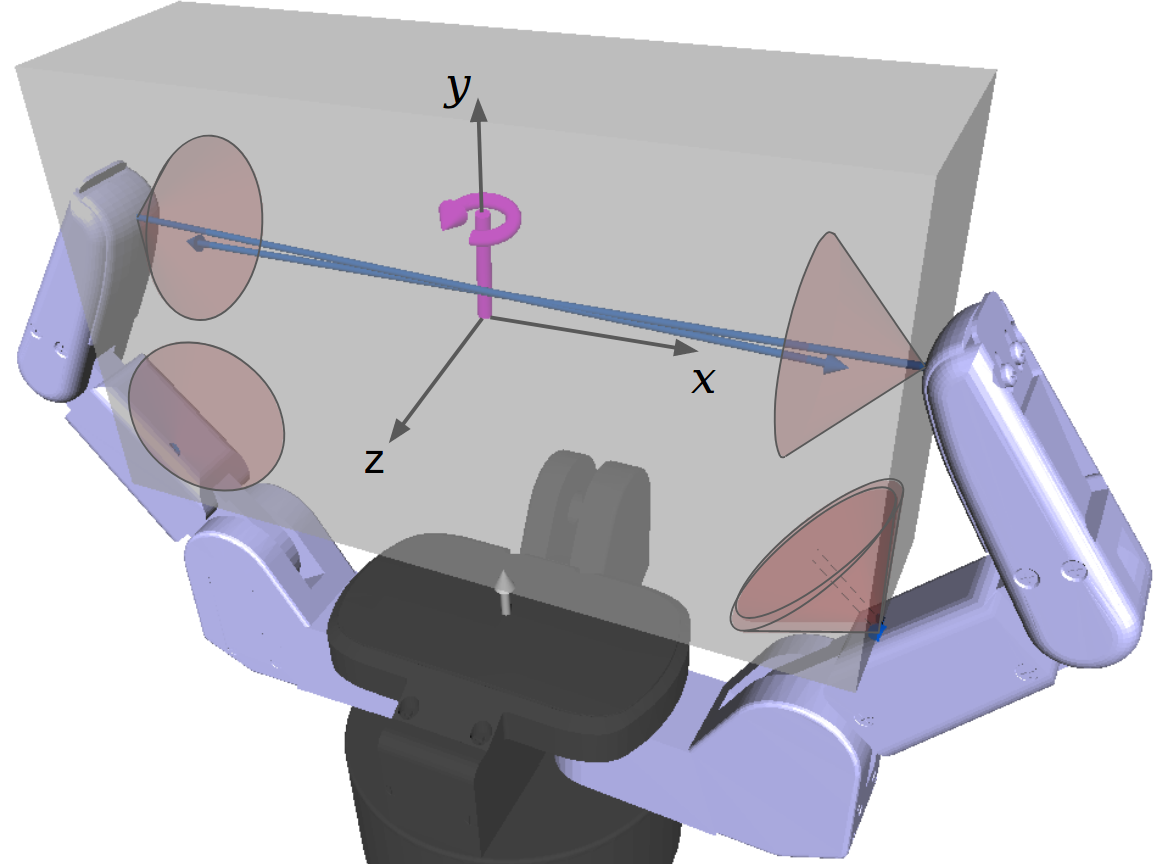}
   \label{fig:unphysical1}} 
\end{subfloat}
\begin{subfloat}[View from above]
   {\includegraphics[width=1.0\columnwidth]{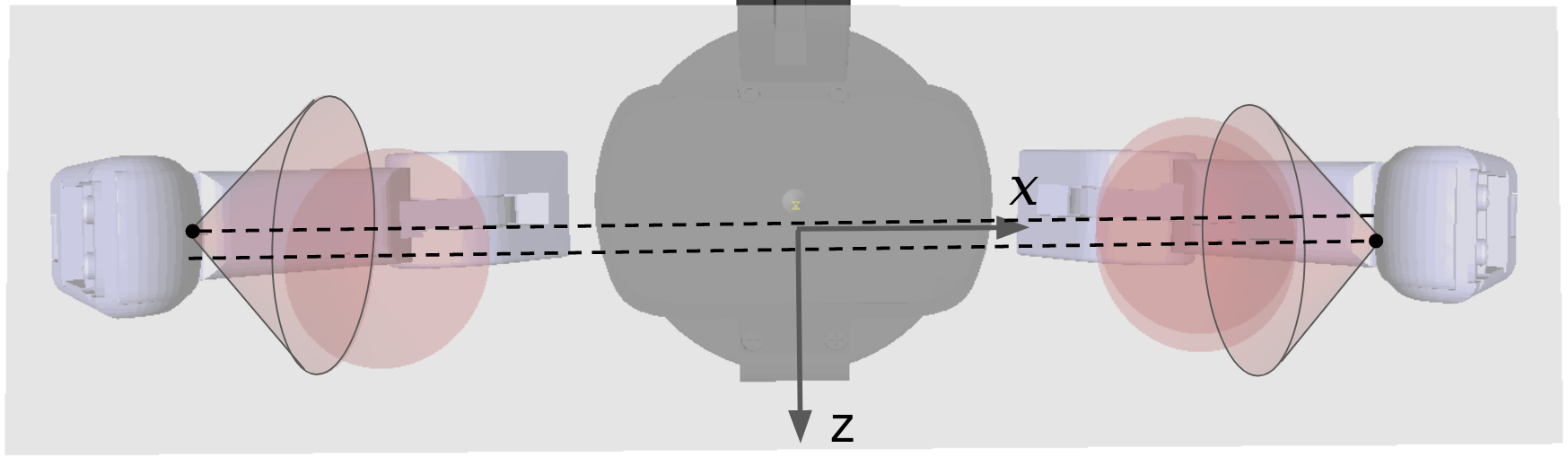}
   \label{fig:unphysical3}}
\end{subfloat}
\caption{Illustration of the shortcomings of neglecting the MDP when solving passive grasp stability problems. A force is applied to the object in the positive $y$-direction. The solver rotates the object around the $y$-axis (violet) generating normal forces (blue) due to the slight offset of the distal contacts. These normal forces are so large the frictional force component is barely noticeable. In this fashion a formulation without the MDP allows the solver to resist arbitrary disturbances through unphysical object motions.}\label{fig:unphysical}
\end{figure}

\section{Results}\label{results}

\subsection{Determining stability and computing contact forces}\label{results_intro}

The simplest query we may make is to determine the stability of a
grasp with respect to a given disturbance. We will use this
application of our method to illustrate the importance of the friction
model we introduce compared to previous literature, and extend to more
complex queries in the following subsections.

Let us solve for the stability of the grasp shown in
Fig.~\ref{fig:package} (and again in Fig.~\ref{fig:unphysical}) when
we apply a force of 1N to the object in the $y$-direction
\textit{without preloading the joints at all.} To answer this query,
we use Algorithm~\ref{alg:refinement} as follows:
\begin{eqnarray} 
\text{objective:}~ & \text{none} \\
\text{additional constraints:} & \bm{w} = [0,1,0,0,0,0]^T \\
& \bm{\tau}_c = 0
\end{eqnarray}
Note that, in the absence of an optimization objective, we are simply
asking if a solution exists that satisfies all the constraints of the
problem, equivalent to determining values for all the unknowns
(contact forces, virtual motions, etc.) such that the grasp is
stabilized. Using Gurobi~\cite{gurobi} as a solver for the constituent
MIPs, Algorithm~\ref{alg:refinement} finds no feasible solution to
this problem (i.e. it predicts the grasp is unstable in the presence
of the given disturbance.) This is the expected result: the grasp may
not resist a force of 1N in the $y$-direction without any preloading,
as it is intuitively clear that the object will slide out.

What happens if, instead of our formulation incorporating the MDP, we
use the friction models commonly used in the grasp force optimization
literature~\cite{SALISBURY83,KERR86,NAKAMURA89,BUSS96,HAN00,BOYD07}?
Together with equations (\ref{eq:wrench_eq})-(\ref{eq:normal_forces})
and (\ref{eq:joint_model}) (which model unilateral contacts and
nonbackdriveable joints) the complete query can then be formulated as
a single MIP. Again using Gurobi as a solver, we find that the
resulting MIP accepts a solution \textit{regardless of the magnitude
  of the applied disturbance}. This is equivalent to saying that the
grasp is stable in such cases even in the absence of any preload,
which is clearly incorrect.

\remindtext{2-4-2] \& [O7-2-5}{Why does a simpler friction model accept solutions that we know are
physically unrealistic? To obtain further insight, we can study the
virtual object motion returned by the solver in such a case: this
motion corresponds to a rotation around the $y$-axis (along which also
the disturbance is applied), ``wedging'' the object between the
rigid, passively loaded fingers (Fig.~\ref{fig:unphysical1}). The
solver exploits a slight offset between contacts
(Fig.~\ref{fig:unphysical3}) to generate very large normal forces,
thus permitting friction forces to balance the applied disturbance,
regardless of its magnitude. However, the equilibrium contact forces
computed in this fashion \textit{do not oppose the virtual motion, and
  thus break the laws of energy conservation}. In order to obtain
physically meaningful results it is necessary to include the MDP in
the friction constraints.}

\begin{figure}[t!]
\centering
\includegraphics[width=1.0\columnwidth]{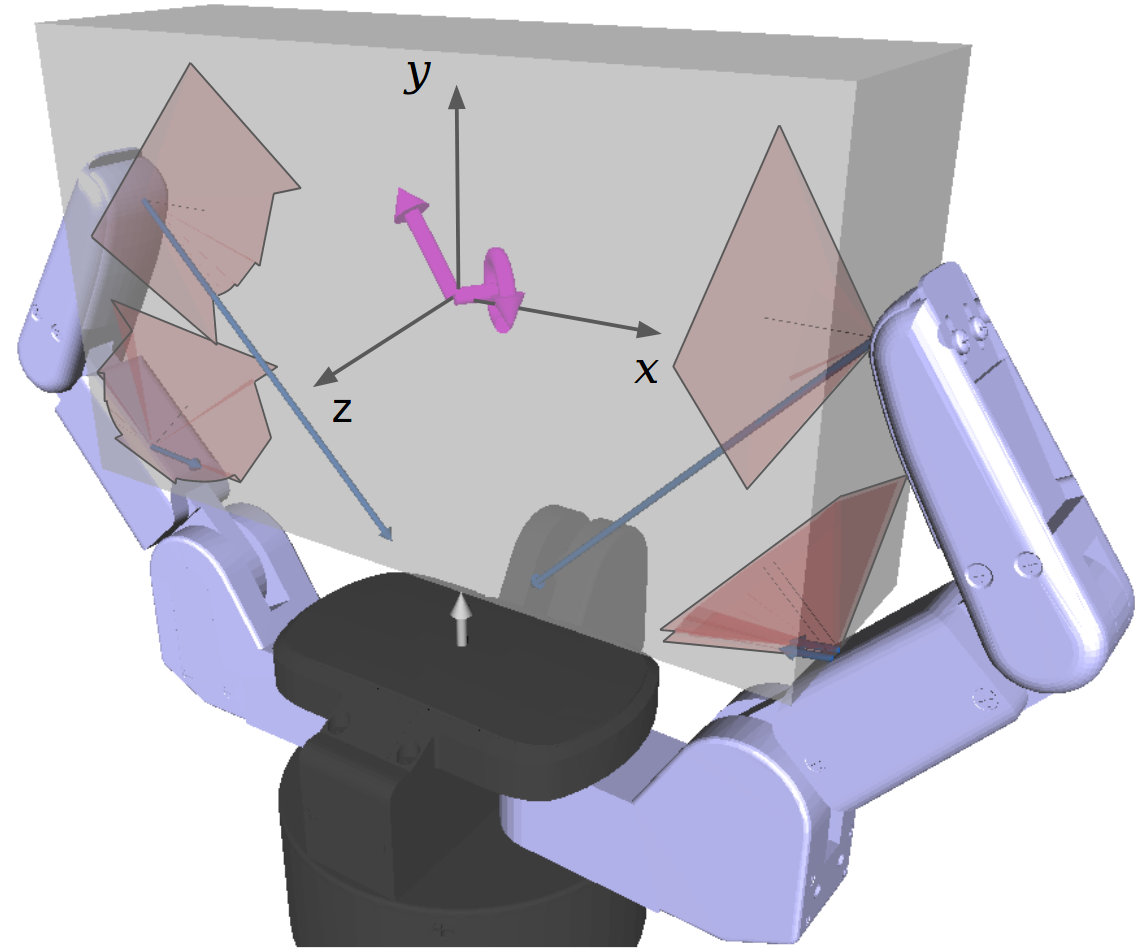}
\caption{Equilibrium contact forces that are predicted to arise by our framework when a force of 2.2N is applied to the object in the $y$-direction. The proximal joints have both been preloaded with 0.1Nm. Note the refinement of the pyramidal friction cone approximation. Only a minority of sectors have been refined. Sectors that are not of interest in this specific grasp problem remain in a less refined state and thus only contribute little to the complexity of the overall problem. The sectors containing the equilibrium contact forces are $<1\degree$~small.}\label{fig:spot_check}
\end{figure}

In contrast, our formulation correctly captures the interplay between applied
preload, and the ability to resist disturbances. Let us now consider the case
where we apply a preload torque of 0.1 Nm at the proximal joints. Due to the
simplicity of the grasp we can analytically determine the expected maximum
force in the $y$-direction the grasp can withstand: a $\sim$90mm contact
moment arm and a friction coefficient of 1.0 results in a maximum total
friction force applied to the object of $\sim$2.2 N. Using a similar
formulation as before (no objective, $\bm{\tau}_c = 0.1$Nm at the proximal
joints), we indeed find that Algorithm 1 accepts a solution (predicts
stability, see Fig.~\ref{fig:spot_check}) for a disturbance of 2.2N in the $y$-direction ($\bm{w} =
[0,2.2,0,0,0,0]^T$), but finds no solution for a disturbance of 2.5N in the
same direction ($\bm{w} = [0,2.5,0,0,0,0]^T$).

\subsection{Maximum resistance to disturbances in a given direction}

\begin{figure}[t!]
\centering
   \begin{subfloat}[No contact normal uncertainty]
   {\includegraphics[width=1.0\columnwidth]{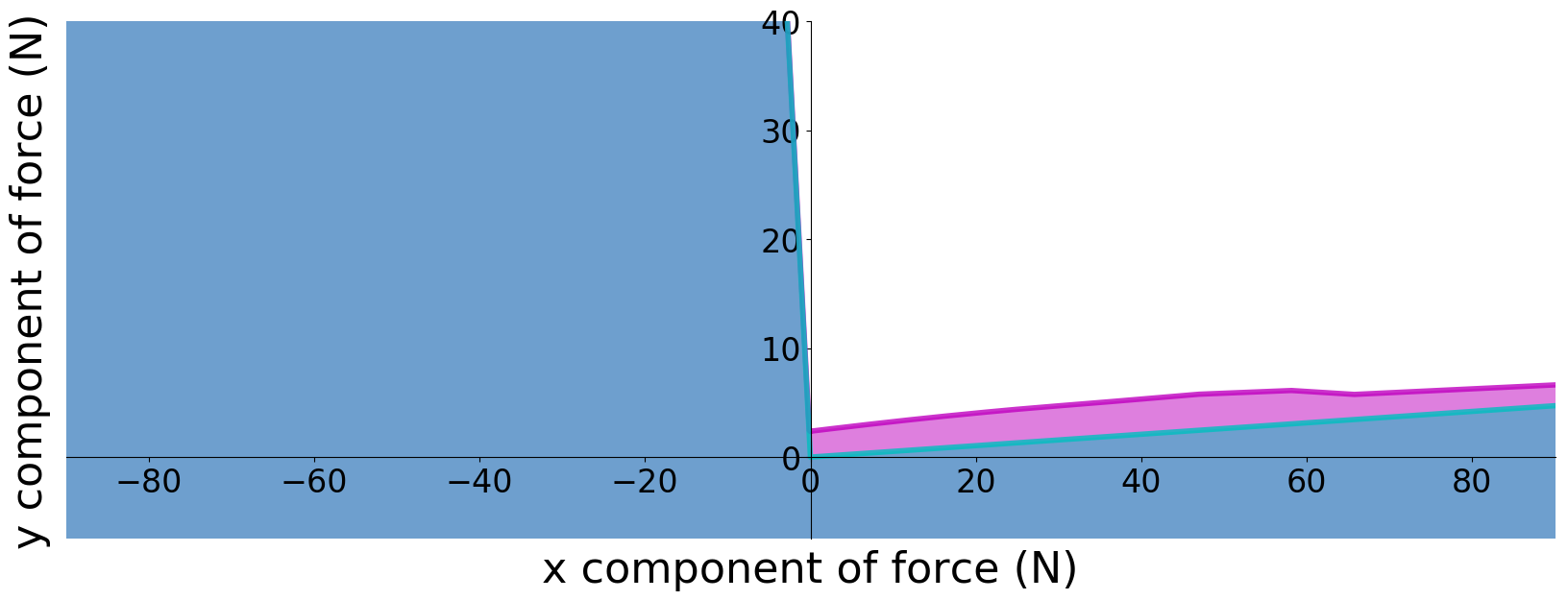}
   \label{fig:package_exact}} 
\end{subfloat}
\begin{subfloat}[Robust to contact normal uncertainties of 2.5\degree]
   {\includegraphics[width=1.0\columnwidth]{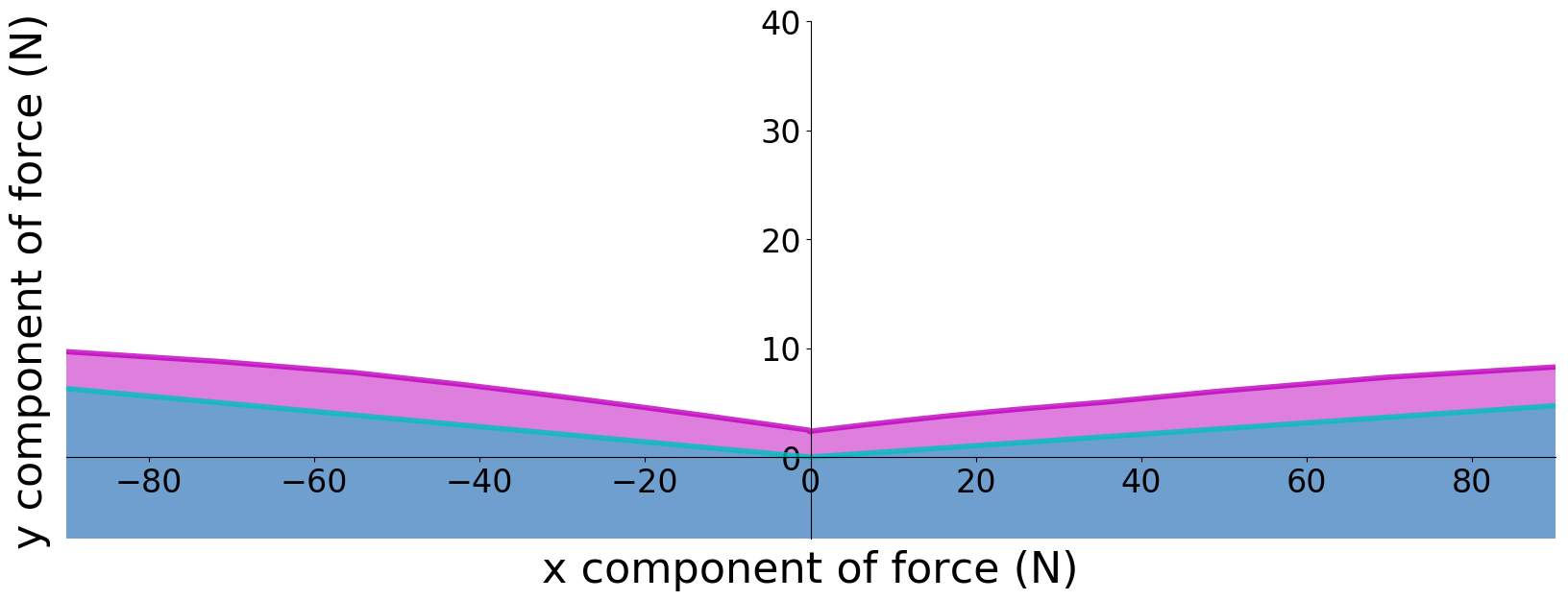}
   \label{fig:package_robust}}
\end{subfloat}
\caption{Resistible forces in the grasp plane of the grasp in Fig.~\ref{fig:package} as predicted by our model and algorithm. In blue are forces that can be resisted even without the application of preloading torques at the joints. When loading the two proximal joints with 0.1Nm the maroon area is added to the resistible forces.}\label{fig:package_map}
\end{figure}

So far we have used Algorithm~\ref{alg:refinement} simply to check
feasibility of a problem, without an optimization objective, and used
it to ``spot check'' resistance to specific disturbances. Adding an
objective allows us to formulate more powerful queries. For example,
we can directly determine the exact \textit{maximum disturbance}
applied to the grasped object in a given direction that a grasp may
resist purely passively.

To achieve this, we prescribe a preload $\bm{\tau}_m$ for the
actuators to be kept constant, and a direction $\bm{d}$ along which to
apply a disturbance to the object. To compute the largest magnitude
disturbance the grasp can withstand in that direction, we use
Algorithm~\ref{alg:refinement} as follows:
\begin{eqnarray} 
\text{objective:}~ & \text{maximize}~ s \\
\text{additional constraints:} & \bm{w} = s\bm{d} \\
& \bm{\tau}_c = \bm{\tau}_m
\end{eqnarray}



To illustrate this, we will again use as an example the grasp in
Fig.~\ref{fig:package}. The simplicity of this grasp allows us to
verify the accuracy of predictions made by our algorithm by comparison
to the the intuitions formulated in Section~\ref{introduction}. We
discretize the space of possible disturbance directions lying in the
$xy$-plane with a 1\degree~resolution. We then compute the maximum
magnitude wrench that can be resisted along each of these discrete
directions. We visualize the results in Fig.~\ref{fig:package_map},
where Fig.~\ref{fig:package_exact} plots results without considering
robustness to contact normal uncertainty, while
Fig.~\ref{fig:package_robust} assumes an uncertainty of 2.5\degree.

The results match our intuition that any downward force can be reacted without
any loading of the fingers. Furthermore the model captures the need for finger
loading in order to resist upward forces. It also shows an effect of passive
finger loading for forces with nonzero $x$ component: pushing sideways
increases the amount of resistance to upwards forces.

\begin{figure}[tbp]
\centering
\includegraphics[width=1.0\columnwidth]{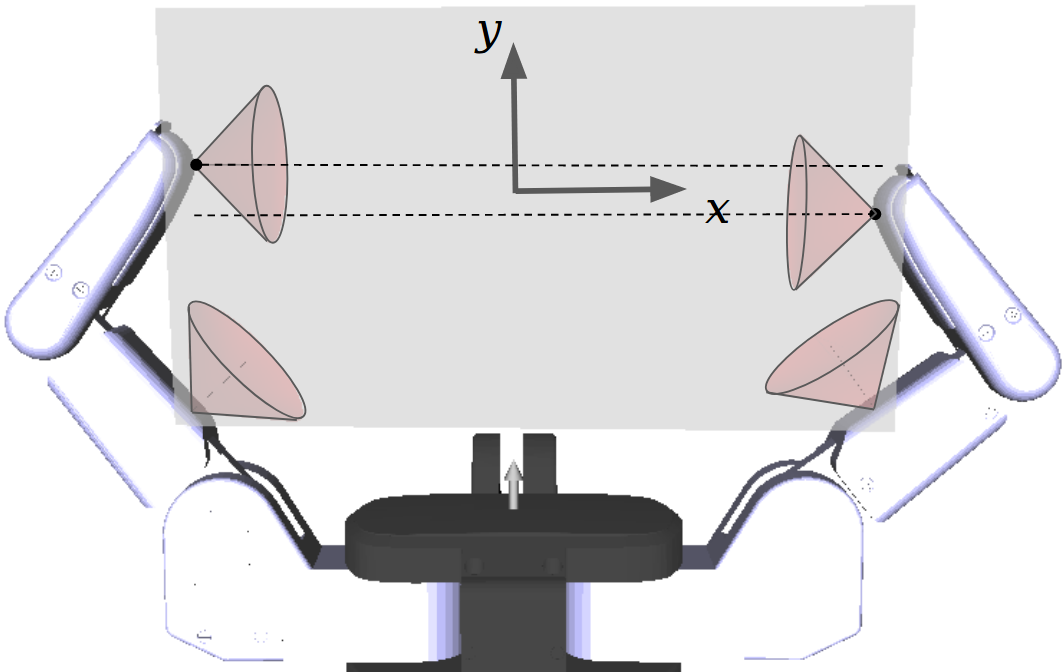}
\caption{A grasp designed to highlight the 'wedging' effect. The two contacts on the distal link are offset with respect to each other allowing wedging to occur.}
\label{fig:skewed_grasp}
\end{figure}

The reason for the asymmetry of Fig.~\ref{fig:package_exact} however
is not immediately obvious, as the grasp itself appears symmetric. In
fact however, the two distal contacts are ever so slightly offset,
causing the object to wedge itself stuck if enough leftward force is
applied. The grasp in Fig.~\ref{fig:skewed_grasp} makes it clearer why
this behavior occurs - here the contacts are visibly offset. Note,
that this wedging behavior is very different from what we observed
when solving without the MDP (see Fig.~\ref{fig:unphysical}). There,
the wedging occurred no matter the applied wrench such that arbitrary
wrenches could be resisted. Furthermore, the resulting contact forces
did not satisfy energy conservation.

In contrast, in our framework only specific wrenches allow wedging to occur.
These wrenches depend on the geometry of the grasp and are consistent with the
rigid body statics of the grasp problem. The equilibrium contact forces
predicted by our framework satisfy the MDP and hence energy conservation.
However, as our method allows us to solve the rigid body problem very
accurately, only a small offset is required for our model to predict wedging
of the object; an offset that is easily within the accuracy of a typical
triangular mesh. Note, that this behavior is fully consistent with the rigid
body assumption and the predictions made by our framework are correct, albeit
highly sensitive to the grasp geometry.

Of course in practice it is not advisable to rely on such volatile
geometric effects. Therefore taking into account geometric
uncertainties is of paramount importance for practical
applications. Fig.~\ref{fig:package_robust} shows which forces can be
robustly resisted when we consider the uncertainty in normal angle to
be no larger than 2.5\degree~(using the approach outlined in
Section~\ref{uncertainties}.) The resulting plot of resistible forces
is approximately symmetric corresponding to the near-symmetry of the
grasp. The indicated spaces of resistible forces both with and without
a preload are consistent with our intuition. \addedtext{7-5}{We note
  that the range of passively resistable forces in the $xy$-plane for
  this grasp was also experimentally validated in previous
  work~\cite{HAASHEGER_TASE18}, and the results shown in
  Fig.~\ref{fig:package_robust} match this existing data.}

\begin{figure}[tbp]
\centering
\includegraphics[width=1.0\columnwidth]{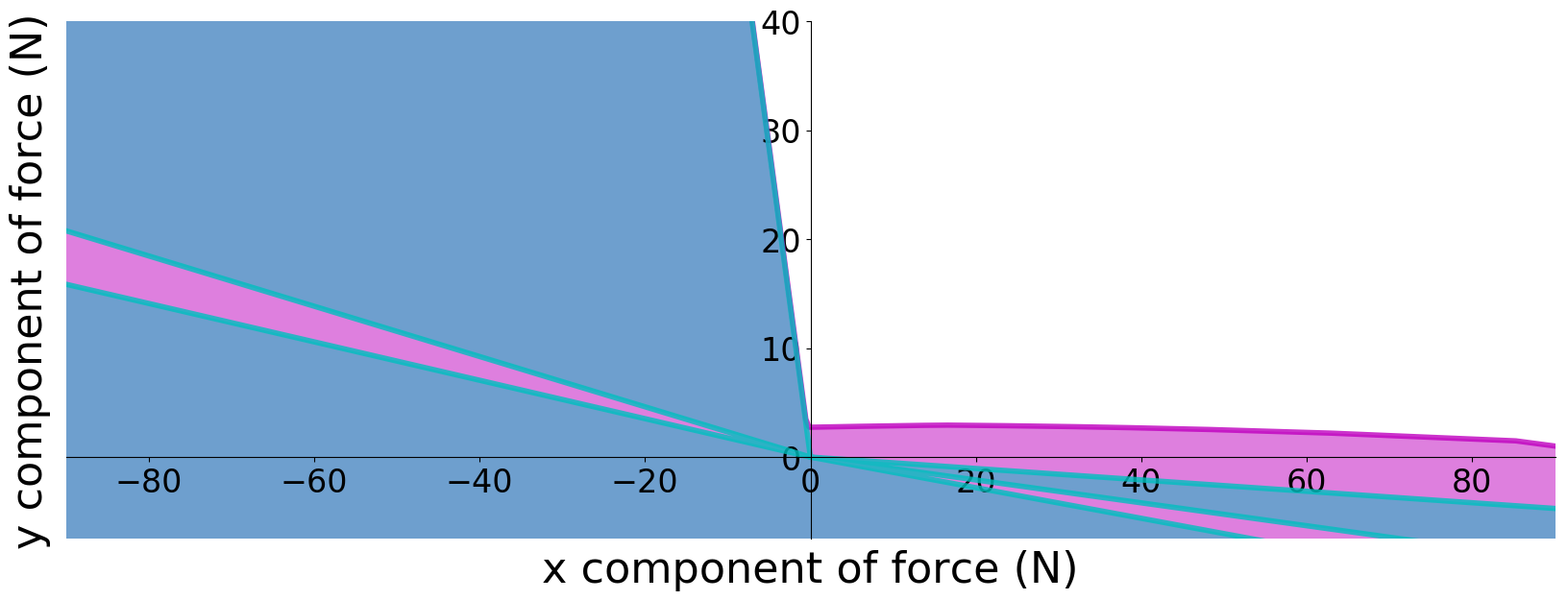}
\caption{Forces in the $xy$-plane predicted resistible by the grasp in Fig.~\ref{fig:skewed_grasp} using our model and algorithm. In blue are forces that can be resisted even without the application of preloading torques at the joints. When loading the two proximal joints with 0.1Nm the maroon area is added to the resistible forces. We also take into account a contact normal uncertainty of 2.5 degrees to make sure the grasp is robust to such discrepancies.}
\label{fig:skewed}
\end{figure}

We showed how we can make the stability predictions less sensitive to the
small scale geometric characteristics of the grasp and thus robust to
uncertainties. At a larger scale, however, wedging effects can be robustly
leveraged. If the contacts are offset such as in Fig.~\ref{fig:skewed_grasp}
the forces this grasp may resist robustly are shown in Fig.~\ref{fig:skewed}.
Thus, if we know the range of disturbances likely to be encountered during a
manipulation task our framework can be a valuable tool in picking an
appropriate grasp.

\addedtext{2-3}{One feature of Fig.~\ref{fig:skewed} perhaps requires further
elaboration: The sectors in the second and fourth quadrants of
Fig.~\ref{fig:skewed} where forces may only be resisted when a preload is
applied but not otherwise. These sectors stand out because they appear thin
and are surrounded by large areas where applied forces are resistible even
without a preload. We thus investigated what effects cause these wedges in
order to verify if these predictions are physically accurate:

\begin{figure}[tbp]
\centerline
{\includegraphics[width=1.0\columnwidth]{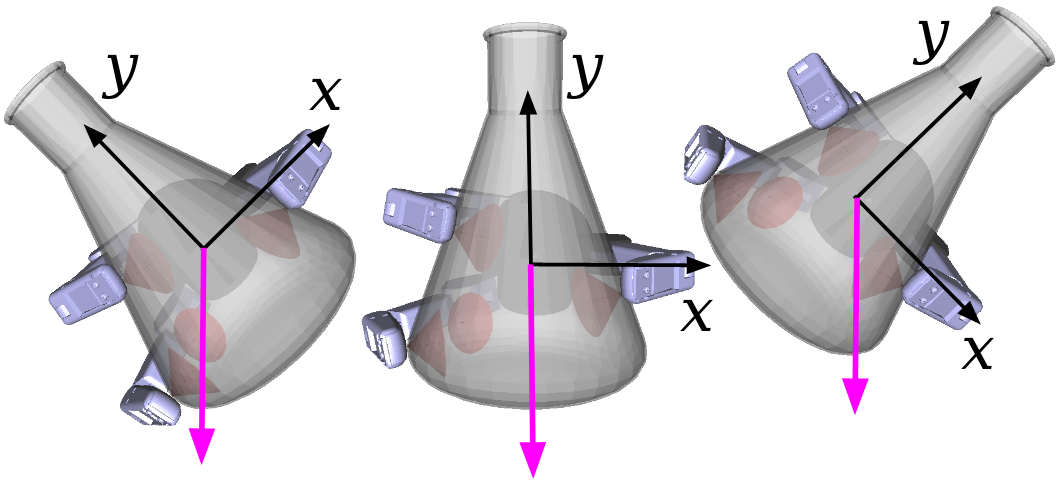}}
\caption{3 dimensional grasp that highlights the necessity of a grasp to be able to withstand a range of forces applied to the object. During a pouring task gravity (pink) moves in the $xy$-plane.}
\label{fig:flasks}
\end{figure}

\begin{figure}[tbp]
\centering
\includegraphics[width=1.0\columnwidth]{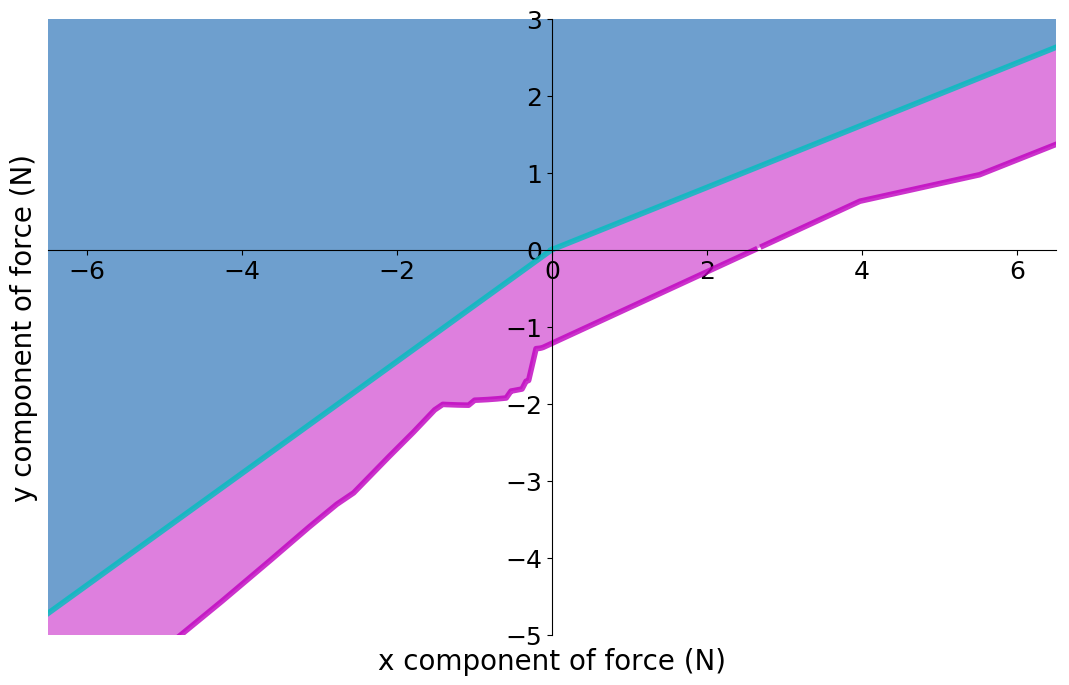}
\caption{Forces in the $xy$-plane predicted resistible by the grasp in Fig.~\ref{fig:flasks}. In blue are forces that can be resisted even without preloading joint torques. When loading the proximal joints with 0.1Nm, the maroon area is added to the resistible forces. We use a contact normal uncertainty of 2.5 degrees.}
\label{fig:flask_map}
\end{figure}

The grasps shown in Fig.~\ref{fig:package} and~\ref{fig:skewed_grasp} were created such that all contacts are as close as possible to lying in a mutual plane. This is because three dimensional grasps can be very complex while two dimensional grasps allow us to use our intuition to validate the predictions made by our framework. A limitation of using GraspIt! in our grasp analysis is that it is difficult to create grasps that are truly two dimensional in nature. Due to the meshing of the finger and object geometries as well as the intricacies of collision checking in GraspIt! the contacts always lie somewhat offset from the central plane.

In the specific case of the grasp in Fig.~\ref{fig:skewed_grasp} the contacts do not quite lie within the $xz$-plane, in which the forces applied to the object lie. This means that two contacts are generally not sufficient in order to balance an applied force. Let us investigate the wedge in the second quadrant of Fig.~\ref{fig:skewed}: When a force is applied along the $(-1,0)$ direction the distal contact of the left finger acts as a fulcrum and the object rotates clockwise loading the contact on the proximal contact of the right finger. When a force is applied in the $(-1,1)$ direction the left finger contact again acts like a fulcrum, however now the object rotates counter-clockwise loading the distal contact on the right finger.

Somewhere in between those two cases the applied force points almost directly at the fulcrum contact and instead of rotating the object is mostly pressed against the left finger breaking both contacts on the right finger. Thus, only two contacts remain and the grasp becomes unstable. In this particular case the object would rotate out of the grasp around the y-axis as the two remaining contacts on the left finger do not lie in the same plane as the applied force.
}

Let us now consider the grasp in Fig.~\ref{fig:flasks}. Note that this
grasp comprises four contacts (one on each distal link plus one on a proximal link)
which do not lie on the same plane, and thus has to be analyzed in a
three-dimensional framework. We consider here an apparent task the
robot grasping the flask may need to execute. In order to pour a
liquid contained in the flask it is necessary to tip it. If we choose
to use the robot wrist for this tipping motion then the force of
gravity acting on the flask and its contents lies in the
$xy$-plane. The grasp must thus be able to resist such forces in order
to complete its task successfully. Furthermore, we have a choice of
direction in which to turn the flask in order to pour its
content. Creating a visualization (shown in Fig.~\ref{fig:flask_map})
as before we can deduce the need for a preload, and that it is more
robust to turn the flask counter-clockwise. Thus, once a grasp has
been established our framework can help in making decisions as to how
a task is to be executed.

\subsection{Computing optimal actuator commands}

\begin{figure}[t]
\centering
\includegraphics[width=0.9\columnwidth]{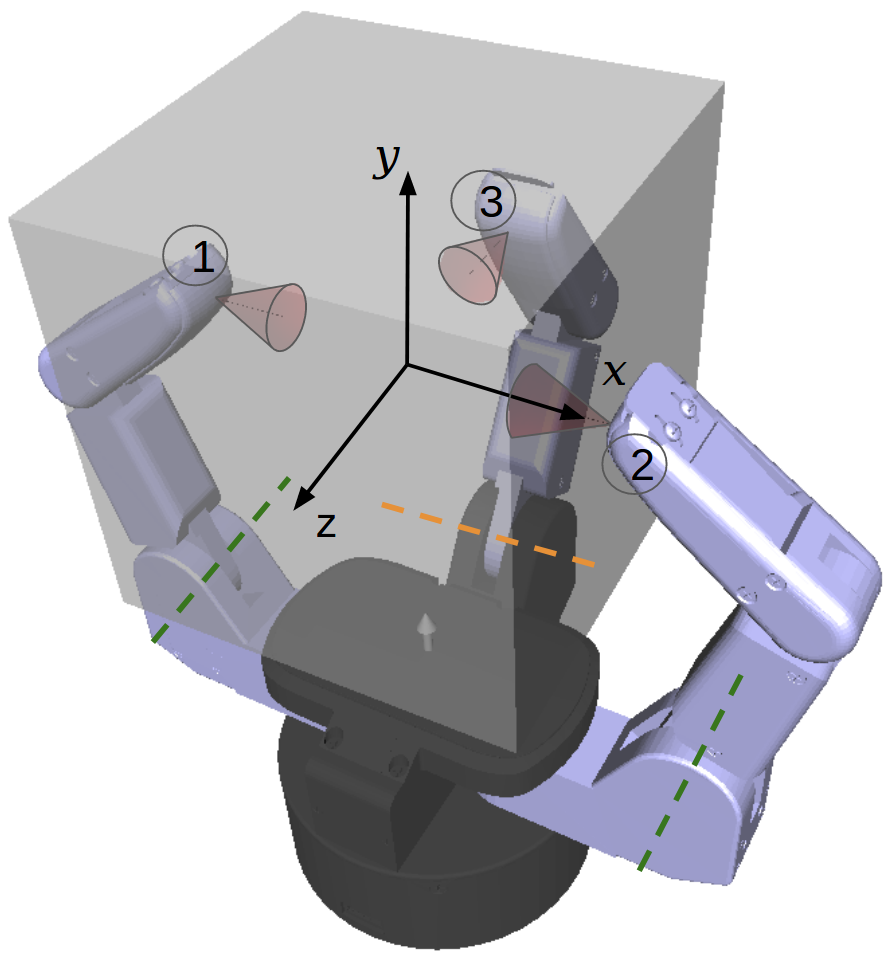}
\caption{Example grasp to illustrate the importance of appropriately choosing grasp preloads. We assume a preload of 0.1Nm at the proximal joints of fingers 1 and 2 (axes marked as green dashed lines). We apply a range of preload torques at the proximal joint of finger 3 (axis marked as yellow dashed line) and evaluate passive stability.}
\label{fig:cube}
\end{figure}

\begin{figure}[t]
\centering
\includegraphics[width=1.0\columnwidth]{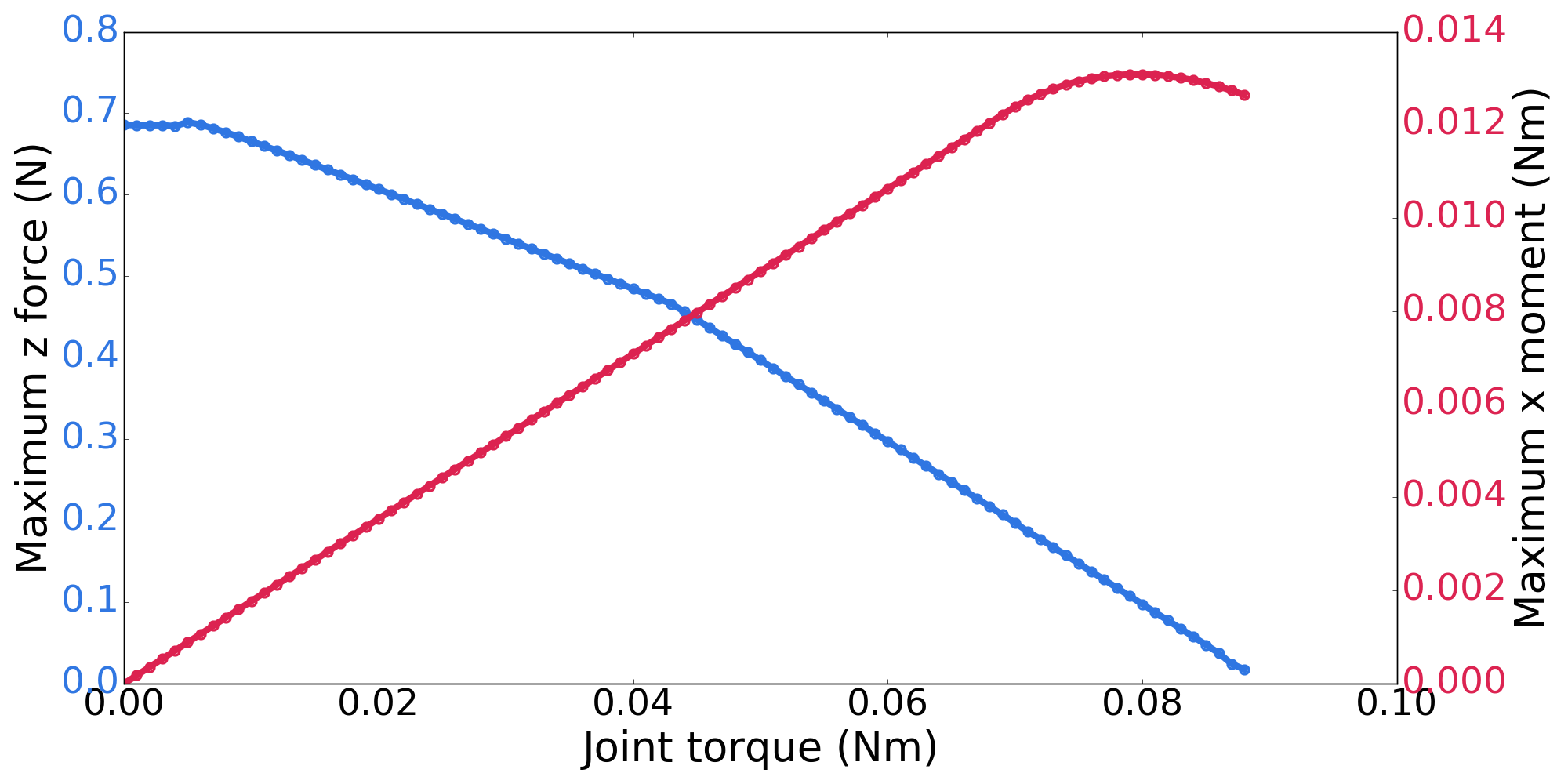}
\caption{Force in the $z$-direction (blue) and moment in the $x$-direction (red) the grasp in Fig.~\ref{fig:cube} can resist for a range of preloads at finger 3.}
\label{fig:preload_test}
\end{figure}

The passive stability of a grasp is not only determined by its geometry: the
actuator commands are equally important. Consider for example the grasp in
Fig.~\ref{fig:cube}. The three contacts the hand makes with the object all lie
approximately in the $xz$-plane. Contacts 1 and 2 lie approximately on the
$x$-axis and oppose each other. Let us assume we create a grasp by commanding
the proximal joints of fingers 1 and 2 to each apply 0.1Nm. Let us vary the
torque commanded at the proximal joint of finger 3 and observe the difference
in passive stability. Specifically, we will use our framework to investigate
the maximum disturbance on the object the grasp can resist in two directions.
Fig.~\ref{fig:preload_test} shows the resulting predictions from our
algorithm.

First we are interested in forces along the positive $z$-axis. As expected,
the resistance is largest if no motor torque is applied by finger 3. Any load
by this finger only adds to the disturbance and does not help in resisting it.
When the torque applied by finger 3 reaches 0.09 Nm, it has completely removed
any resistance to $z$-direction forces. This can be easily verified: The
coefficient of friction chosen for this example is 0.45 and the moment arms
from joint to contact are identical for all three fingers. As there cannot be
any out-of-plane forces, the normal forces at all contacts will be
proportional to the applied joint torque. Thus, applying 0.9Nm at finger 3
claims all possible contact friction at both contacts 1 and 2: no further
forces in that direction can be resisted.

Let us now consider passive resistance to torques applied to the
object around the $x$-axis. If we do not load finger 3 the object is
only held by contacts 1 and 2. As both these contacts lie on the
$x$-axis they cannot apply any torque to the object in that
direction. Thus, the grasp cannot resist any torques around the
$x$-axis unless we also load finger 3. The third finger provides the
contact necessary for resisting the torque on the object. The more we
load finger 3, the larger the force at contact 3 and the larger the
resistible torque. At some point however, as discussed above, the
forces at finger 3 begin to overwhelm fingers 1 and 2 and the object
slides out along the $z$ axis even without any external disturbances.

\begin{table*}[t]
\centering
\begin{tabular}{P{0.1\textwidth} | P{0.1\textwidth} P{0.1\textwidth} P{0.15\textwidth} | P{0.1\textwidth} P{0.1\textwidth} P{0.15\textwidth} }
\multirow{2}{0.1\textwidth}{Equivalent number of friction edges} & \multicolumn{3}{c|}{\textbf{Without robustness scheme}} & \multicolumn{3}{c}{\textbf{With robustness scheme}} \\
& Maximum wrench (N) & Time at full resolution (s) & Time with hierarchical refinement (s) & Maximum wrench (N) & Time at full resolution (s) & Time with hierarchical refinement (s) \\
\hline
4 & \textgreater 100 & 0.088 & 0.088 & \textgreater 100 & 0.11 & 0.11 \\
8 & 7.07435 & 0.22 & 0.43 & 7.07435 & 0.24 & 0.46 \\
16 & 6.34905 & 0.43 & 0.65 & 3.73377 & 0.45 & 1.02 \\
32 & 3.71036 & 0.72 & 1.92 & 3.70539 & 0.97 & 1.33 \\
64 & 3.70095 & 1.58 & 2.33 & 2.31479 & 1.83 & 3.09 \\
128 & 3.69775 & 5.80 & 3.50 & 1.21805 & 5.12 & 10.9 \\
256 & 3.69751 & \textgreater 600 & 4.87 & 1.21756 & 20.3 & 13.0 \\
512 & 3.69681 & - & 4.34 & 1.21743 & 102 & 17.8 \\
1024 & 3.07268 & - & 13.3 & 1.21740 & 394 & 22.9 \\
2048 & 3.07267 & - & 15.0 & 1.21739 & - & 32.2 
\end{tabular}
\caption{Performance of the analysis of the grasp in Fig.~\ref{fig:flasks}. We compute the magnitude of the largest force that can be applied in the negative $y$-direction without destabilizing the grasp. We do this for varying levels of refinement expressed as the number of sectors the friction cone approximation contains. We record compute times for both an approach where all sectors are already of the desired size and our hierarchical refinement approach.}
\label{tab:data}
\end{table*}

This example shows the importance of preload and passive stability:
for this grasp, loading finger 3 helps resistance against some
disturbances, but hurts against others. The right amount of preload
must thus be chosen based on the task. To the best of our knowledge,
no existing grasp stability analysis method can make such predictions.

Using our model, we can also find actuator commands that are optimal
with respect to any specific objective we chose. For example we may want to
minimize the maximum torque a single actuator must produce to resist a given
wrench $\bm{w}_m$. We now use Algorithm~\ref{alg:refinement} as follows:
\begin{eqnarray} 
\text{objective:}~ & \text{minimize}~ \max_{j}{\tau_c^j} \\
\text{additional constraints:}~ & \bm{w} = \bm{w}_m 
\end{eqnarray}

For the grasp in Fig.~\ref{fig:flasks}, we can compute the optimal actuator
commands for a force of ($\bm{w} = [0,-1.21743,0,0,0,0]^T$), the largest force in
the negative $y$-direction that can be resisted when applying a preload of
0.1Nm at every proximal joint, according to our previous analysis. We find
that the optimal torques at these joints are actually only ($\bm{\tau}_c = [0,
0.042, 0.073]^T$). This shows that a large amount of the preload (0.1Nm at
every joint) is wasted in the sense that it does not increase disturbance
resistance in this particular direction.

\addedtext{10-3}{In many practical applications it may be of interest to take
into account physical limits such as the maximum torque an actuator can apply
or a maximum permissible normal force in order to not break the grasped
object. Such constraints can be expressed as linear inequalities and are
straightforward to add to our model.}

\begin{figure*}[t!]
\centering
\subfloat[]{\includegraphics[width=1.15in]{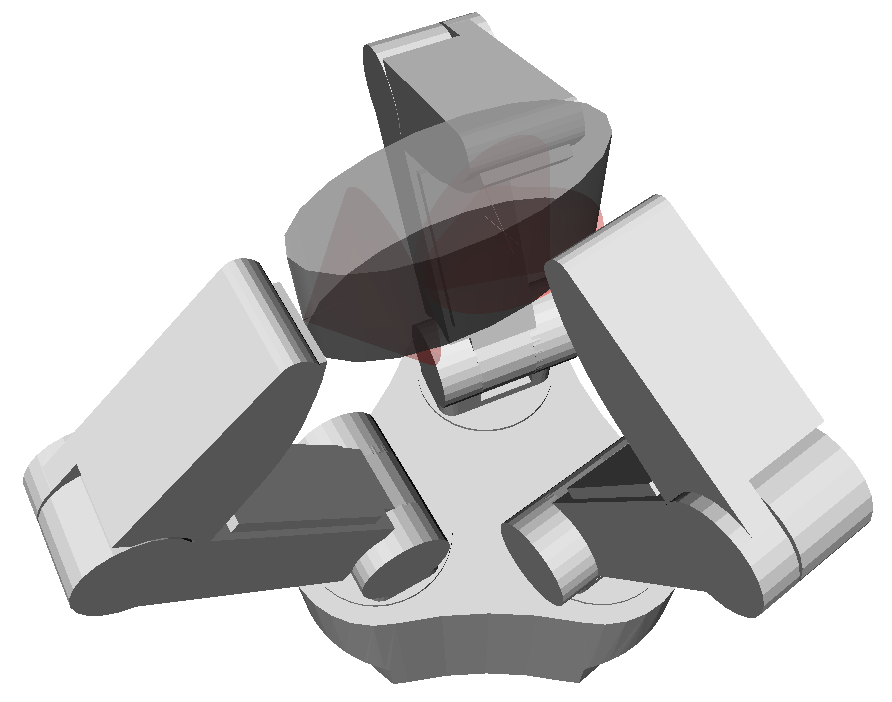}}\hfil%
\subfloat[]{\includegraphics[width=1.15in]{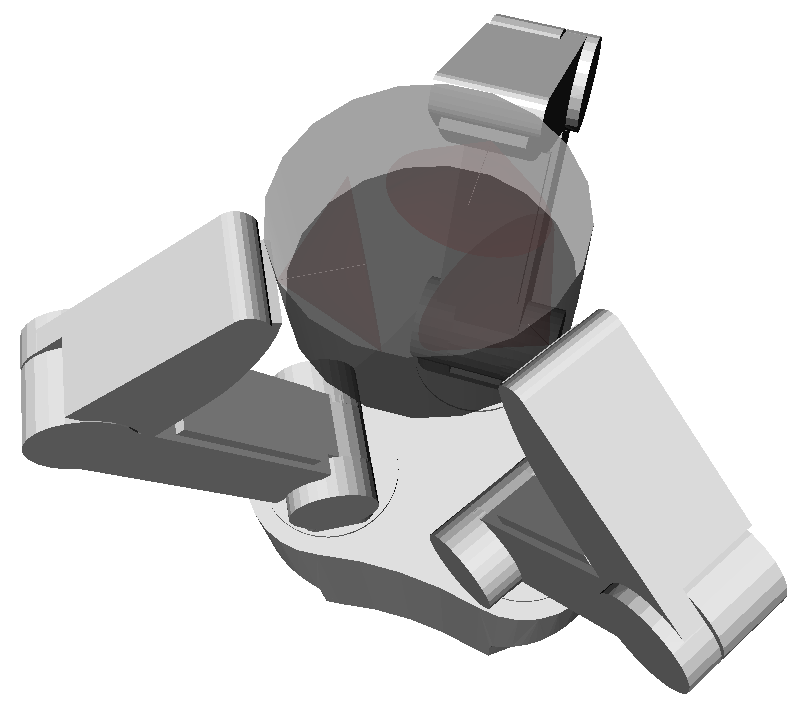}}\hfil%
\subfloat[]{\includegraphics[width=1.15in]{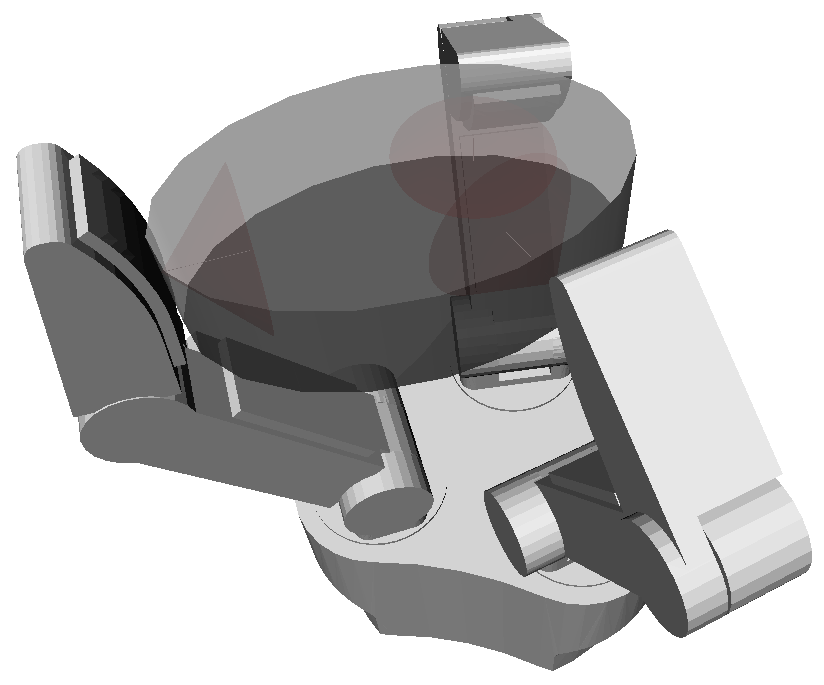}}\hfil%
\subfloat[]{\includegraphics[width=1.15in]{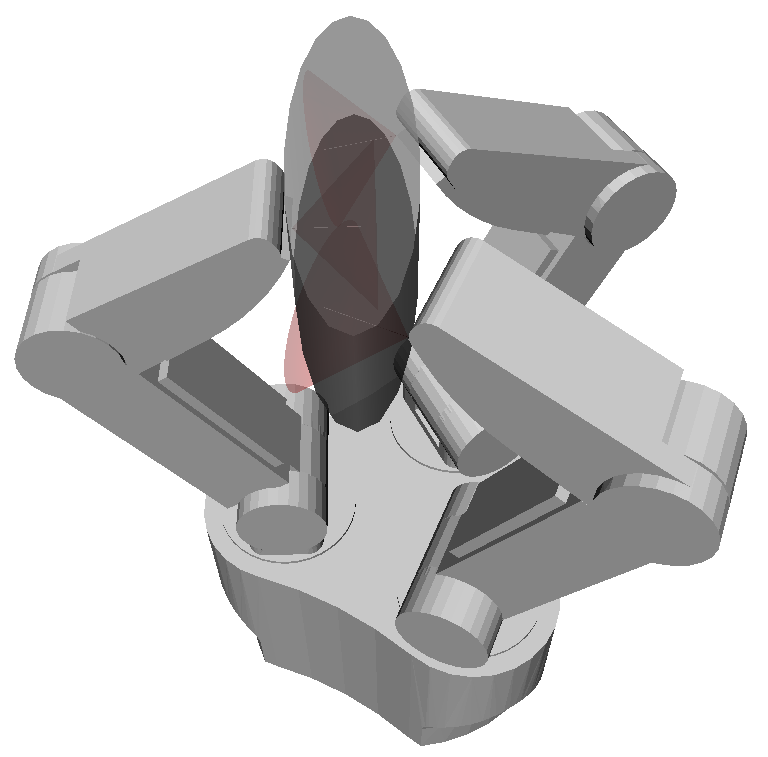}}\hfil%
\subfloat[]{\includegraphics[width=1.15in]{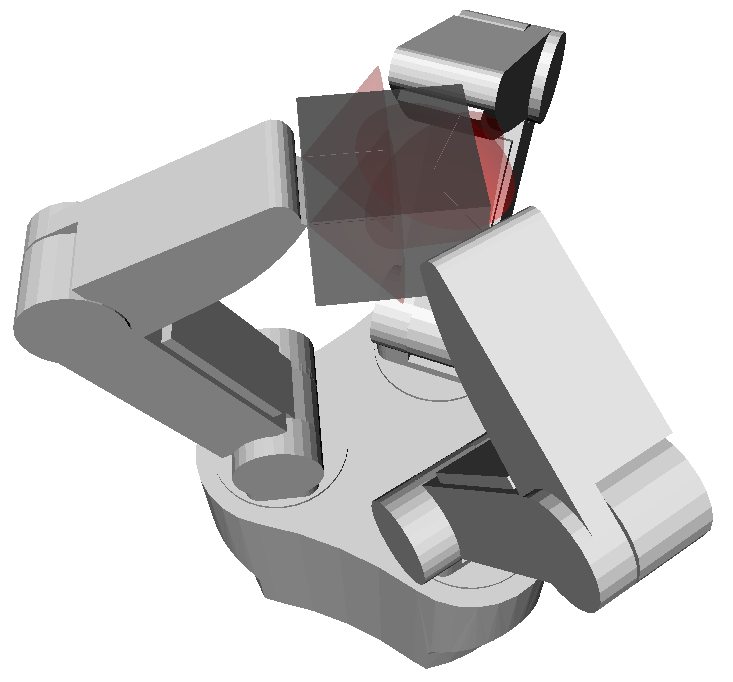}}\hfil%
\subfloat[]{\includegraphics[width=1.15in]{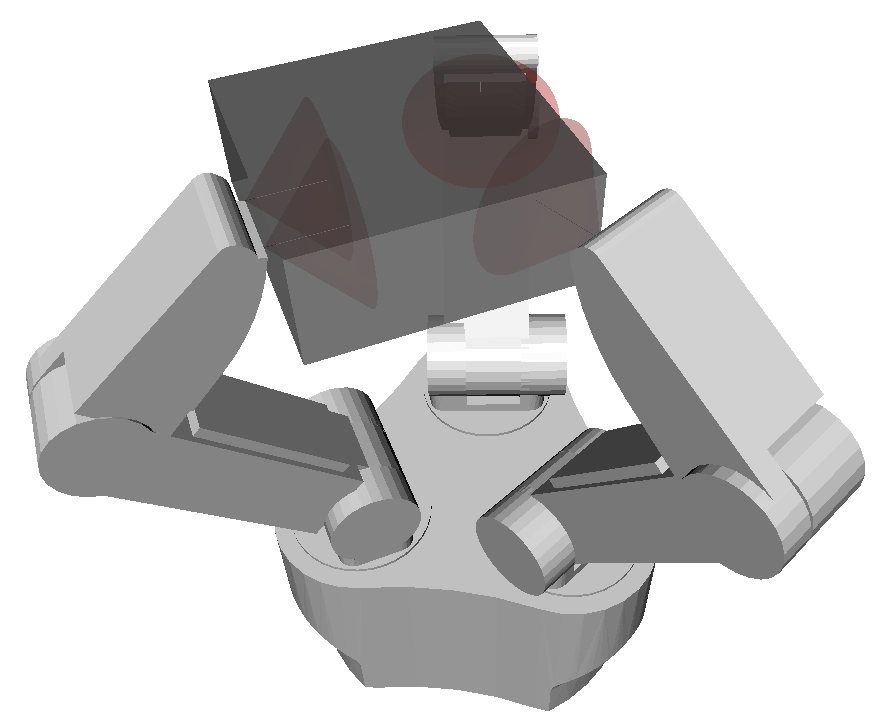}}\hfil%
\subfloat[]{\includegraphics[width=1.15in]{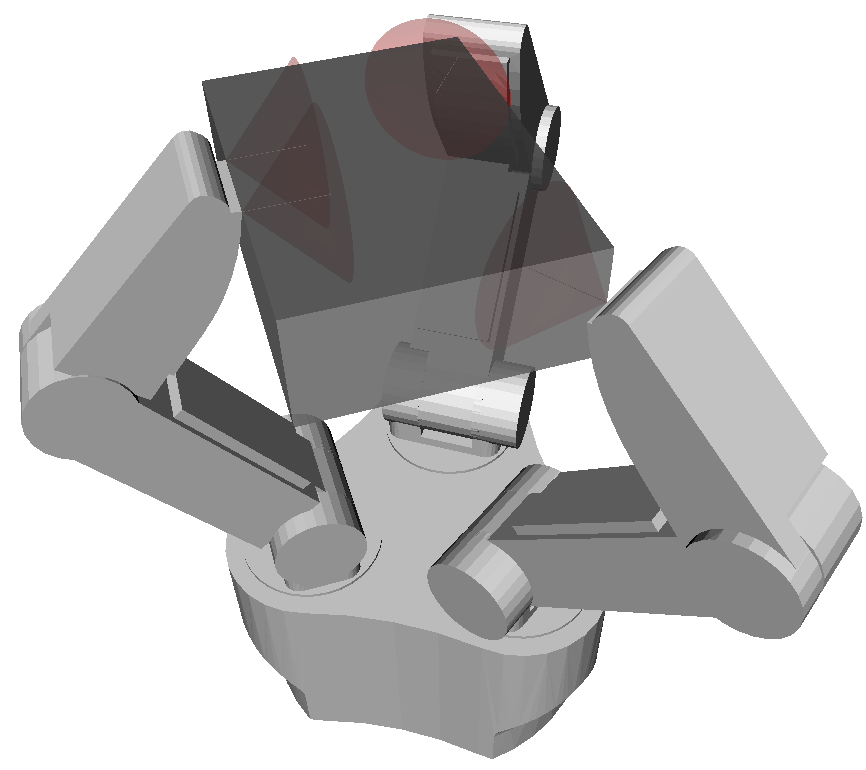}}\hfil%
\subfloat[]{\includegraphics[width=1.15in]{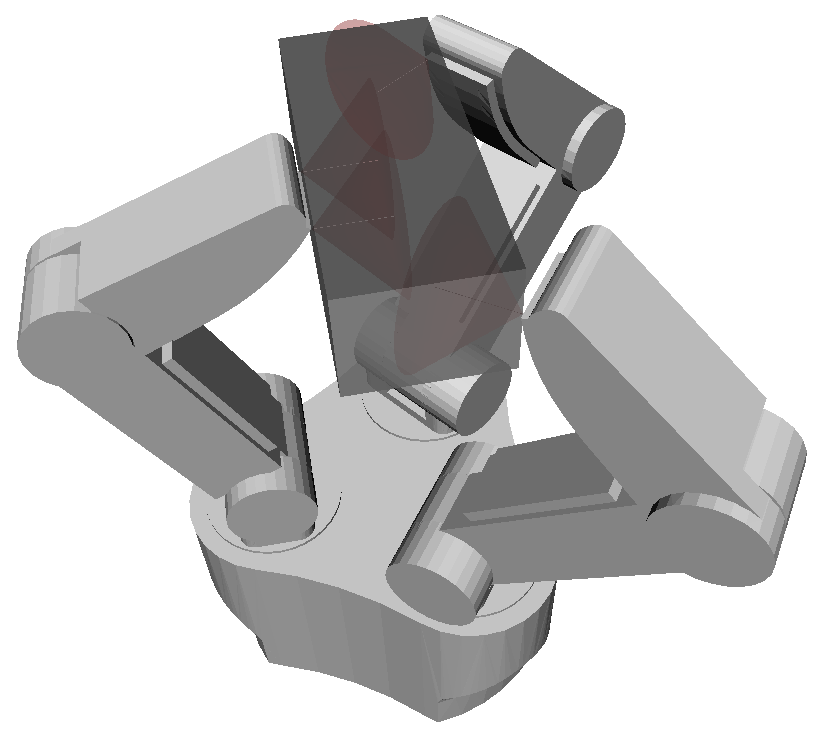}}\hfil%
\subfloat[]{\includegraphics[width=1.15in]{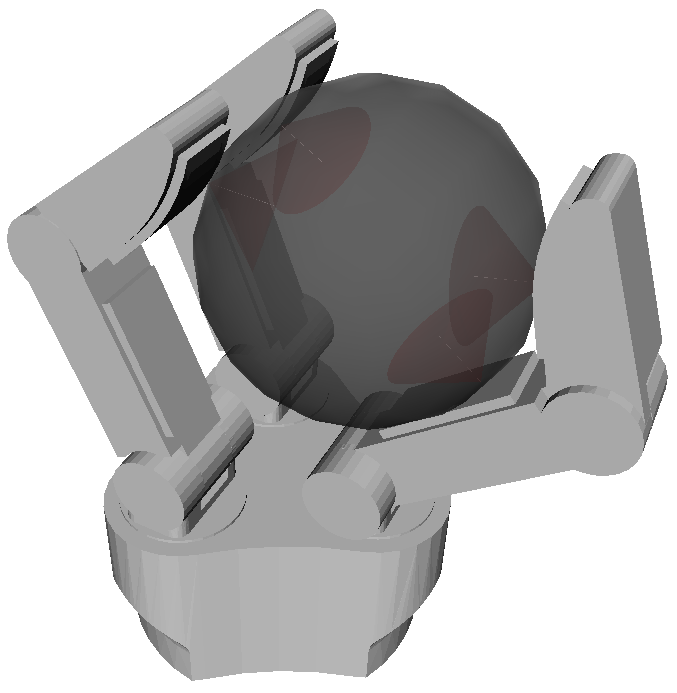}}\hfil%
\subfloat[]{\includegraphics[width=1.15in]{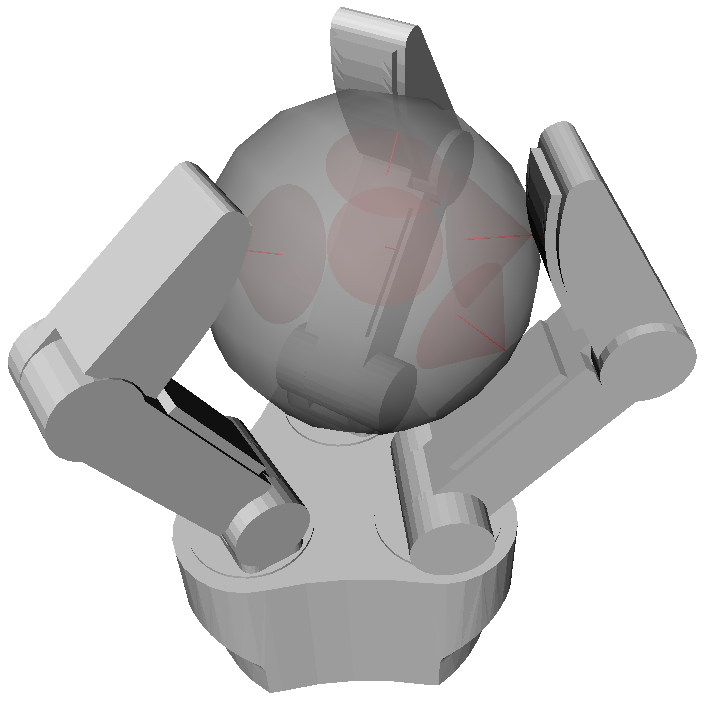}}\hfil%
\subfloat[]{\includegraphics[width=1.15in]{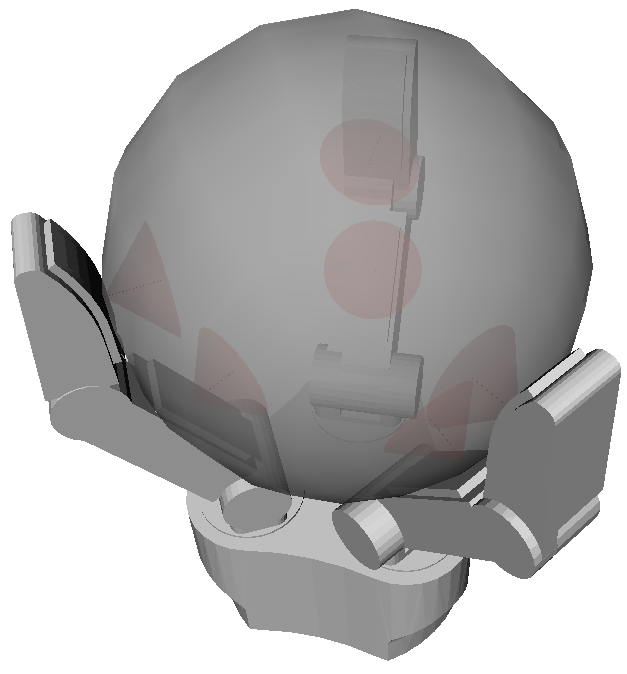}}\hfil%
\subfloat[]{\includegraphics[width=1.15in]{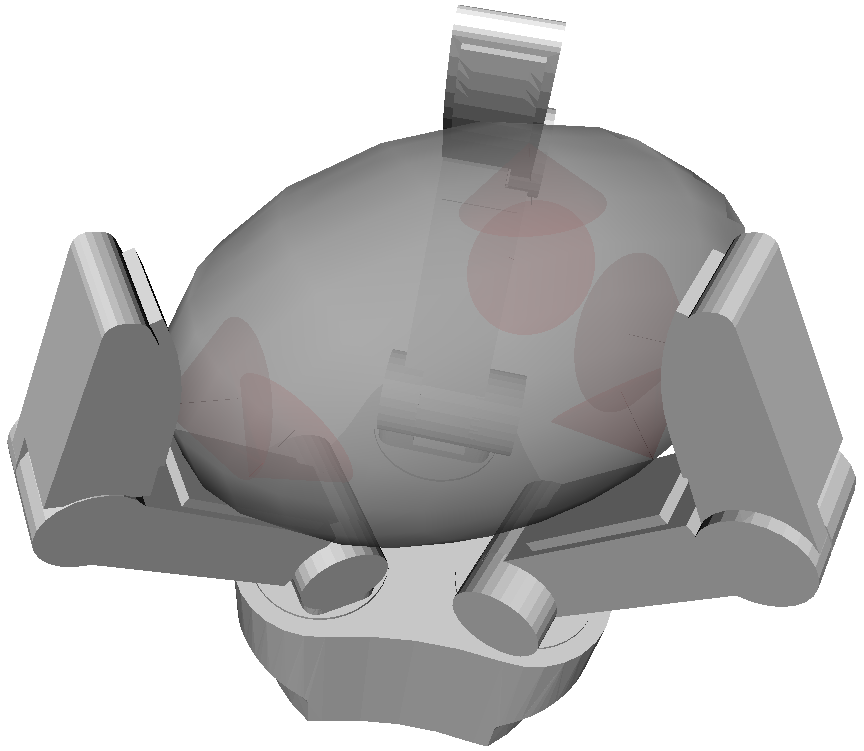}}%
\caption{\addedtext{7-3-1}{Algorithmically generated grasps to investigate the computational performance of our method.}}\label{fig:comp_grasps}
\end{figure*}

\begin{table*}[t]
\centering
\begin{tabular}{P{0.05\textwidth} P{0.075\textwidth} | c | c c | c c}
\multirow{2}{0.1\textwidth}{\textbf{Grasp}} & \multirow{2}{0.1\textwidth}{\textbf{Number of contacts}} & \textbf{Stability check} & \multicolumn{2}{c|}{\textbf{Maximum disturbance}} & \multicolumn{2}{c}{\textbf{Optimal torques}} \\
& & Time (s) & Mean time (s) & Median time (s) & Mean time (s) & Median time (s) \\
\hline
(a) & 3 & 0.246 & 0.797 $\pm$ 0.067 & 0.754 & 0.389 $\pm$ 0.063 & 0.343 \\
(b) & 3 & 0.234 & 0.785 $\pm$ 0.085 & 0.745 & 0.447 $\pm$ 0.074 & 0.408 \\
(c) & 3 & 0.294 & 0.893 $\pm$ 0.056 & 0.907 & 0.347 $\pm$ 0.017 & 0.345 \\
(d) & 3 & 0.259 & 0.664 $\pm$ 0.081 & 0.560 & 0.308 $\pm$ 0.036 & 0.281 \\
(e) & 4 & 1.07 & 10.0 $\pm$ 3.4~ & 6.39 & 3.26 $\pm$ 0.50 & 3.83 \\
(f) & 4 & 15.7 & 36.7 $\pm$ 6.5~ & 39.5 & 13.5 $\pm$ 4.8~ & 9.58 \\
(g) & 4 & 2.78 & 33.8 $\pm$ 9.3~ & 20.8 & 16.8 $\pm$ 5.3~ & 12.3 \\
(h) & 4 & 8.02 & 23.7 $\pm$ 4.2~ & 23.2 & 12.3 $\pm$ 3.3~ & 8.10 \\
(i) & 4 & 1.66 & 14.4 $\pm$ 2.7~ & 11.4 & 1.67 $\pm$ 0.12 & 1.60 \\
(j) & 5 & 6.06 & 150 $\pm$ 24~ & 105 & 22.8 $\pm$ 5.8~ & 19.0 \\
(k) & 6 & 51.0 & 398 $\pm$ 104 & 344 & 136 $\pm$ 54~ & 80.6 \\
(l) & 6 & 27.5 & 488 $\pm$ 99~ & 330 & 777 $\pm$ 288 & 626 \\
\end{tabular}

\caption{\addedtext{7-3-2}{Runtime analysis of our method for the three tasks
demonstrated in this paper performed on a consumer desktop computer for the
grasps shown in Fig.~\ref{fig:comp_grasps}. Where multiple trials were
performed we report the mean $\pm$ standard error as well as the median
runtimes.}}\label{tab:comp_data}

\end{table*}

\subsection{Analysis of computational performance}

\addedtext{7-3-3}{In Section~\ref{refinement} we state that an
  accurate solution to the grasp problem with discretized friction
  cones but without hierarchical refinement requires a very large
  number of friction edges. We further argue that solving such a
  problem becomes computationally intractable as a large number of
  friction edges results in a large number of binary variables in the
  MIP.} To verify these hypotheses, we analyze the convergence of our
algorithm with varying levels of refinement of the friction
approximation. We use the grasp in Fig.~\ref{fig:flasks} as an
example, with the task of finding the largest force in the negative
$x$-direction that the grasp can withstand.  We do so at varying
levels of refinement, and record the predicted force magnitude and
runtime.

We compare two approaches: The first method ("full resolution") directly uses
a friction cone approximation that is at the desired level of accuracy in its
entirety. The second method ("hierarchical refinement") always starts with a
coarse approximation and refines as described in
Algorithm~\ref{alg:refinement}. Throughout all experiments, both methods (when
able to finish)  produced identical solutions, but the running times varied
greatly. All recorded data can be found in Table~\ref{tab:data}.

We notice that, at high levels of refinement, full resolution becomes
intractable, whereas hierarchical refinement finds a solution
efficiently. The study of how the refinement level affects the
returned solution is more complex. The exact value of the solution
generally reaches a point where increasing the accuracy of the
approximation (adding more friction edges) stops making a significant
difference. In some cases, as in the case of the maximum wrench in the
left side of Table~\ref{tab:data}, this happens for accuracy levels
that only hierarchical refinement can reach. In others, as in the case
of the maximum robust wrench (with 2.5\degree~normal uncertainty) in
the right side of Table~\ref{tab:data}, both methods are able to find
good approximations of the final value. At the more shallow levels,
full resolution will often outperform hierarchical refinement, but
since we generally do not know which of these cases any specific query
might fall into, only hierarchical resolution allows us to increase
the accuracy without the risk of compute time
exploding. \addedtext{7-3-4}{These results show both that high
accuracy is actually required in order to obtain meaningful results,
and that solving such problems without hierarchical refinement
is computationally intractable.}

\addedtext{2-4-3}{Another interesting finding is that the predicted maximum
resistible force is exceedingly large when only four friction edges are used
in the first step of the refinement process. This is because at this stage the
MDP is only enforced such that the friction force and negative relative
tangential contact motion lie within a 90\degree~sector. This allows
sufficient freedom to the solver to use the rigidity of the robot hand when
backdriven along with unphysical object motions to create large contact forces
- much like what was described in Section~\ref{results_intro} where there are
no constraints on the friction direction at all. This illustrates again the
need for a high accuracy solution to the grasp problem including the MDP.}

\addedtext{7-3-4}{In order to investigate the practical applicability of our
method, we tested the runtime of our algorithm on a range of grasps that
could be encountered in a grasping task (see Fig.~\ref{fig:comp_grasps}.) We
generated these grasps on a range of differently shaped objects using a brute
force grasp planner~\cite{MEEKER19}. On each of these grasps we applied our
framework to perform the three tasks demonstrated above: Checking for stability,
finding the maximum resistible wrench in a given direction and computing the
optimal actuator commands. In the second task the direction along which to
find the maximum resistible disturbance can have an impact on the runtime of
the algorithm. We hence repeated the solution process with ten different
randomly generated directions to obtain more meaningful results. Similarlty,
the time it takes to optimize the actuator commands depends on the external
wrench applied to the object and hence we also repeat these experiments with
ten randomly sampled external wrenches. The results are recorded in
Table~\ref{tab:comp_data}. In order to guarantee accurate results we continued
refinement until we reached a level equivalent to 2048 sectors in the friction
approximation. Empirically we have found this level of refinement guarantees
convergence for all grasps tested (although many converged sooner.)

We note that, as expected, the runtime of our algorithm grows with the
number of contacts. For grasps with three contacts, all queries
typically had sub-second runtimes; up to and including five contacts
we noticed runtimes typically between 1 and 30 seconds. Also as
expected, stability checks (the must fundamental operation needed for
a grasp) are faster than computations like maximum resistable
disturbance or optimal joint torques. This suggests different
applications of our framework for different scenarios: pruning a
larger number of possible grasps using the faster stability analysis,
then computing the optimal torques only on the most promising
candidates. Even with six contacts, our method had a runtime on the
order of minutes, suitable for example for fixturing analysis in a
manufacturing line. Finally, we also note that, in the absence of our
hierarchical refinement, it is altogether intractable to approach this
level of accuracy for all except the simplest of grasps.}

\section{CONCLUSIONS}

In this paper, we have described a grasp stability model that allows for
efficient and accurate solution methods under realistic constraints. Noting
that an exact formulation of Coulomb friction includes non-convex constraints
(due to the Maximum Dissipation Principle), we use a discretization method
that allows the problem to be reformulated as a piecewise convex Mixed Integer
Program solvable through branch and bound. However, such discretization
methods traditionally involve a trade-off: coarse discretizations provide only
rough approximations of the exact constraints, while high resolutions
discretizations are computationally intractable.

To address this problem, we introduce a hierarchical refinement
method that progressively increases the resolution of the
discretization only in the relevant areas, guided by the solution
found at coarser levels. Our local refinement method remains efficient
up to high discretization resolution, and also provides strong
guarantees: if a solution can not be found at a coarse approximation
level, the underlying exact problem is guaranteed not to have a
solution either. Combined, these two features make our method
efficient for problems both with and without exact solutions. It is,
to the best of our knowledge, the first time that grasp stability
models incorporating Coulomb friction (along with the MDP) have been
solved with such high discretization resolution.

Our motivating factor for including the MDP in our grasp stability
analysis is to study what we refer to as passive grasp stability. This
phenomenon occurs when torques at a hand's nonbackdrivable joints
arise passively in response to a disturbance, allowing the grasp to
resist the disturbance without the need for any actively applied motor
torque. This is a ubiquitous situation in practice, and one that
greatly expands the usability of many robot hands equipped with highly
geared motors. However, no current grasp model can distinguish between
the disturbances that will be resisted by passive effects and those
that will not.

We thus combine our friction model with models of unilateral contacts
and nonbackdrivable joints (also formulated as Mixed Integer
constraints). The overall model accepts many types of queries: for
example, we can analyze the space of wrenches applied to an object
that a given grasp can withstand, or compute optimal joint commands
given a specific object wrench. Thanks to the hierarchical refinement
method, these can be solved efficiently (on the order of seconds per
query) even with very high resolution approximations of the MDP.

Running our analysis method on a number of example grasps, we showed
that our method predicts effects both intuitive (pressing directly
against contacts is passively stable, but pulling the object away
requires preload torques) and more subtle (an object wedging itself in
a grasp in response to a disturbance for a given contact geometry.) In
contrast, grasp stability models that do not consider the MDP produce
unrealistic results, and fail to predict the dependence of disturbance
resistance on applied preloads.

A limitation of our method is that, while it performs well in
practice, its theoretical running time remains worst-case exponential
in the level of discretization for friction constraints. Furthermore,
for cases where a coarse discretization yields a sufficiently accurate
solution, hierarchical refinement might be outperformed by an
equivalent method with uniform resolution (although these cases are
generally unknowable in advance, without actually solving up to high
resolutions.)

\addedtext{10-5}{A further limitation is an arguably narrow definition of
grasp stability. An initially unstable grasp may - through movement of of the
fingers and object and hence the contact - eventually settle in a different
stable equilibrium grasp. In such a case our framework can only determine that
the initial grasp is unstable and makes no prediction on stability of the
final grasp. In order to account for the motion of an initially unstable grasp
we would have to model the dynamics of the grasp. As discussed in
Section~\ref{model} using the currently available dynamics engines to this end
comes with its own difficulties.}

From a practical perspective, in future work we would like to explore
additional applications of our approach to grasp analysis and planning.
\addedtext{10-4-4}{While the runtime of our algorithm currently is too large
for online grasp planning with more than a few contacts we see the main
practical relevance of our method in enabling practitioners to understand and
utilize passive effects in grasping and in providing labels for learning-based
grasp planners.} The model we introduced can allow multiple types of queries,
and in this paper we have only presented some of the possible applications.
Furthermore, we would like to extend our own previous work on the passive
stability of underactuated hands~\cite{HAASHEGER_TASE18}. Passive stability
phenomena are an important feature of underactuation and we believe the
framework presented in this paper is well suited to the investigation of such
problems. 

We believe the framework may also be applicable to problems encountered in the
field of robotic locomotion, for instance to determine the balance of a legged
robot on uneven terrain~\cite{BRETL08}. From a theoretical perspective we aim
to study the possibility of deriving an algorithm with similar applicability
and a guaranteed polynomial running time in all cases. In previous work we
achieved this for planar grasps~\cite{HAASHEGER_RSS18} and believe this work
can provide insights into how to develop a computationally efficient algorithm
for three dimensional grasping.

Implementations of the grasp model and algorithm presented in this paper are
publicly available as part of the open source GraspIt!
simulator~\cite{MILLER04}.

\bibliographystyle{plainnat}
\bibliography{bib/grasping,bib/thesis}

\begin{IEEEbiography}[{\includegraphics[width=1in,height=1.25in,clip,keepaspectratio]{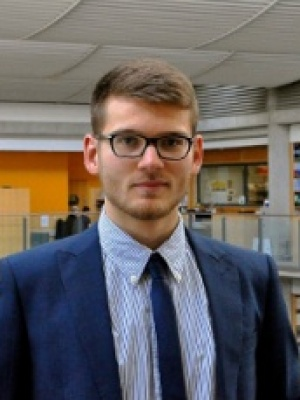}}]{Maximilian Haas-Heger}
Maximilian Haas-Heger received the MEng degree in Aeronautical Engineering from Imperial College London in 2015. Since 2015, he is a PhD candidate in the Robotic Manipulation and Mobility Lab at Columbia University in New York. His research focuses on the theoretical foundations of robotic grasping; specifically on the development of accurate grasp models and their application for dexterous manipulation. 
\end{IEEEbiography}

\begin{IEEEbiography}[{\includegraphics[width=1in,height=1.25in,clip,keepaspectratio]{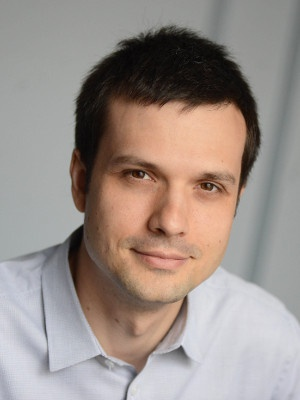}}]{Matei Ciocarlie}
Matei Ciocarlie (S'07-M'12) earned the Ph.D. degree in Computer Science from Columbia University, New York, NY, USA. He was a Research Scientist at Willow Garage, Inc., Menlo Park, CA, USA, and a Senior Research Scientist with Google, Inc., Mountain View, CA, USA. He is currently an Associate Professor of Mechanical Engineering with affiliated appointments in the Computer Science Department and Data Science Institute at Columbia University, New York, NY, USA. His work focuses on robot motor control, mechanism and sensor design, planning and learning, all aiming to demonstrate complex motor skills such as dexterous manipulation.
\end{IEEEbiography}

\end{document}